\pdfoutput=1

\documentclass[11pt]{article}
\usepackage{float}
\usepackage{acl}

\usepackage{times}
\usepackage{latexsym}
\usepackage[T4,T1]{fontenc}

\usepackage[utf8]{inputenc}

\usepackage{microtype}

\usepackage{inconsolata}

\usepackage{times}
\usepackage{latexsym}
\usepackage{tabularx}
\usepackage{graphicx}
\usepackage{todonotes}
\usepackage{makecell}
\usepackage{amsmath}
\usepackage{amssymb}
\usepackage{covington}
\usepackage{subcaption}
\usepackage{tipa}
\title{Mapping `when'-clauses in Latin American and Caribbean languages: \\an experiment in subtoken-based typology}

\author{Nilo Pedrazzini \\
  The Alan Turing Institute (London, United Kingdom) \\
  \texttt{npedrazzini@turing.ac.uk}}

\begin{document}
\maketitle
\begin{abstract}
Languages can encode temporal subordination lexically, via subordinating conjunctions, and morphologically, by marking the relation on the predicate. Systematic cross-linguistic variation among the former can be studied using well-established token-based typological approaches to token-aligned parallel corpora. Variation among different morphological means is instead much harder to tackle and therefore more poorly understood, despite being predominant in several language groups. This paper explores variation in the expression of generic temporal subordination (`when'-clauses) among the languages of Latin America and the Caribbean, where morphological marking is particularly common. It presents probabilistic semantic maps computed on the basis of the languages of the region, thus avoiding bias towards the many world's languages that exclusively use lexified connectors, incorporating associations between character $n$-grams and English \textit{when}. The approach allows capturing morphological clause-linkage devices in addition to lexified connectors, paving the way for larger-scale, strategy-agnostic analyses of typological variation in temporal subordination.
\end{abstract}


\section{Introduction}\label{sec:intro}

Across the 7000+ world's languages recorded by the Glottolog database (\citealt{nordhoffhammarstrom}, \citealt{glottolog2021})\footnote{\url{https://glottolog.org}} there is great variation in how temporal relations between different eventualities can be encoded in a sentence or discourse unit. English has one main generic temporal subordinator, \textit{when}, which is relatively underspecified with respect to the temporal semantic relation between the clause it introduces and its matrix clause, compared to semantically more precise connectors (e.g. \textit{after}, \textit{before}, or \textit{while}). The number and scope of generic temporal subordinators can vary cross-linguistically from one (e.g. Italian \textit{quando}), to two (e.g. German \textit{wenn}/\textit{als}) or several more (e.g. Pular \textit{nde}/\textit{si}/\textit{\textipa{\!b}ay}/\textit{fewndo}/\textit{tuma}; \citealt{pulargrammar}; \citealt{anon2}). Crucially, languages can additionally or exclusively encode \textsc{when}-clauses\footnote{Small caps \textsc{when} is used to refer to the semantic concept of `generic temporal subordination', rather than the English lemma \textit{when} (written in italics).} morphologically on the predicate, rather than using a lexified subordinator (cf. Spanish \textit{viendo} `see.\textsc{ger}'\footnote{The following abbreviations are used in glosses throughout this paper: \textsc{ger} = gerund, \textsc{3} = third person, \textsc{sg} = singular, \textsc{pl} = plural, \textsc{sbj} = subject, \textsc{vis} = visible (speaker's area), \textsc{ss} = same subject, \textsc{ds} = different subject, \textsc{distr} = distributive, \textsc{narr} = narrative, \textsc{nsbj} = non-subject, \textsc{loc} = locative, \textsc{as2} = secondary assertion, \textit{pro} = prominent, \textsc{pfv} = perfective.} as opposed to \textit{cuando vio} `when saw.\textsc{3.sg}'; Ukrainian \textit{pobačyvšy} `see.\textsc{ger}' as opposed to \textit{koly vyn pobačyv} `when he saw'). Because of the very nature of competition, overarching semantic differences between subordination strategies within individual languages cannot be fully captured in terms of discrete, categorical variables, but they should be modeled as a continuum allowing for a degree of overlap, aiming to reveal broader patterns in a probabilistic, rather than a fully deterministic way. Previous studies (\citeauthor{anon1} \citeyear{anon1}) have employed a `token-based approach' (\citealt{levshina19,levshina21}) to explore the semantic ground covered by English \textit{when} and \textit{induce} cross-linguistically common semantic dimensions from parallel corpora. In \citeauthor{anon1} (\citeyear{anon1}), probabilistic semantic maps (\citealt{croft-poole-2008,walchli-cysouw2012}) of \textsc{when} were generated from a massively parallel corpus of 1400+ linguistic varieties (\citealt{mayer-cysouw}), to capture systematic variation in the ways languages tend to divide the semantic space of English \textsc{when} by using different lexical items for its different meanings.  One of the greatest limitations of a purely token-based typological approach to the study of temporal subordination in the world's languages is that it does not allow to account for variation \textit{within} the semantic space covered by non-lexified \textsc{when}-clauses cross-linguistically. That is, it will merely allow us to observe that particular \textit{subsets} of \textit{when}-occurrences are more likely to lack a parallel token in the target languages, without further identifying typologically widespread constructions (or \textit{gram types}; \citealt{dahl}) within the semantic sub-space of non-lexified \textsc{when}-clauses. 

While languages using predominantly or exclusively morphological means to express generic temporal subordination are relatively uncommon among European languages, non-lexified \textsc{when}-clauses are instead particularly frequent among Latin American languages, as evidenced by the plethora of areal studies on converbal, clause-bridging, and, especially, switch-reference morphology in the region (among others, \citealt{gijnetalsouthame,vangijn2012southame,vangijn2016srsouthame,overall2014,overall2016}). 

This paper zooms in on the languages of Latin American and the Caribbean, given the particular computational challenges posed by their common, extensive use of non-lexified \textsc{when}-clauses (exclusively so or in addition to lexified means). As in previous experiments, \citeauthor{mayer-cysouw}'s (\citeyear{mayer-cysouw}) massively parallel corpus of New Testament translations is used, and probabilistic semantic maps are adopted as a base method to induce typologically relevant dimensions within the semantic space of \textsc{when}, since they allow capturing the gradience and overlap between different means in any given language, as well as the language-internal variation which is inherent to the very concept of competition. The goal of this paper is twofold:
\begin{itemize}
    \item[a.] incorporate associations between character $n$-grams and English \textit{when} for capturing differences among \textsc{when}-clauses that are expressed morphologically \textit{as well as} lexically, and generate probabilistic semantic maps based on the parallel dataset thus refined. As detailed in Section \ref{sec:method}, this method builds on \citeauthor{superpivot}'s (\citeyear{superpivot}) `SuperPivot' approach, but with substantial changes to their pipeline. Crucially, it gets rid of the assumption that there should be at most one `pivot' (i.e. a marker in a parallel language) per linguistic feature (e.g. `past' in \citeauthor{superpivot}'s \citeyear{superpivot} example), reflecting instead the existing typological knowledge about the nature of generic temporal subordination as a phenomenon with great language-internal variation. The code to achieve this is released alongside this paper as a generalized tool, which starts from one or several lexical items in a source language and can be used to look for systematic cross-linguistic variation in a parallel dataset, both at the lexical and morphological level;
    \item[b.] generate probabilistic semantic maps that are built exclusively on the basis of the languages of the region, thus avoiding bias towards the many world's languages that exclusively or predominantly use lexified connectors. The resulting maps and parallel data enriched with $n$-gram annotation are also released to facilitate future computational experiments.\footnote{The code, datasets and all the maps, only a very small portion of which is presented in this paper, can be found in the \href{https://doi.org/10.6084/m9.figshare.25431814.v1}{associated repository}.}
\end{itemize}

\section{Methods}\label{sec:method}

\begin{figure}[!h]
  \centering
  \includegraphics[scale=0.40]{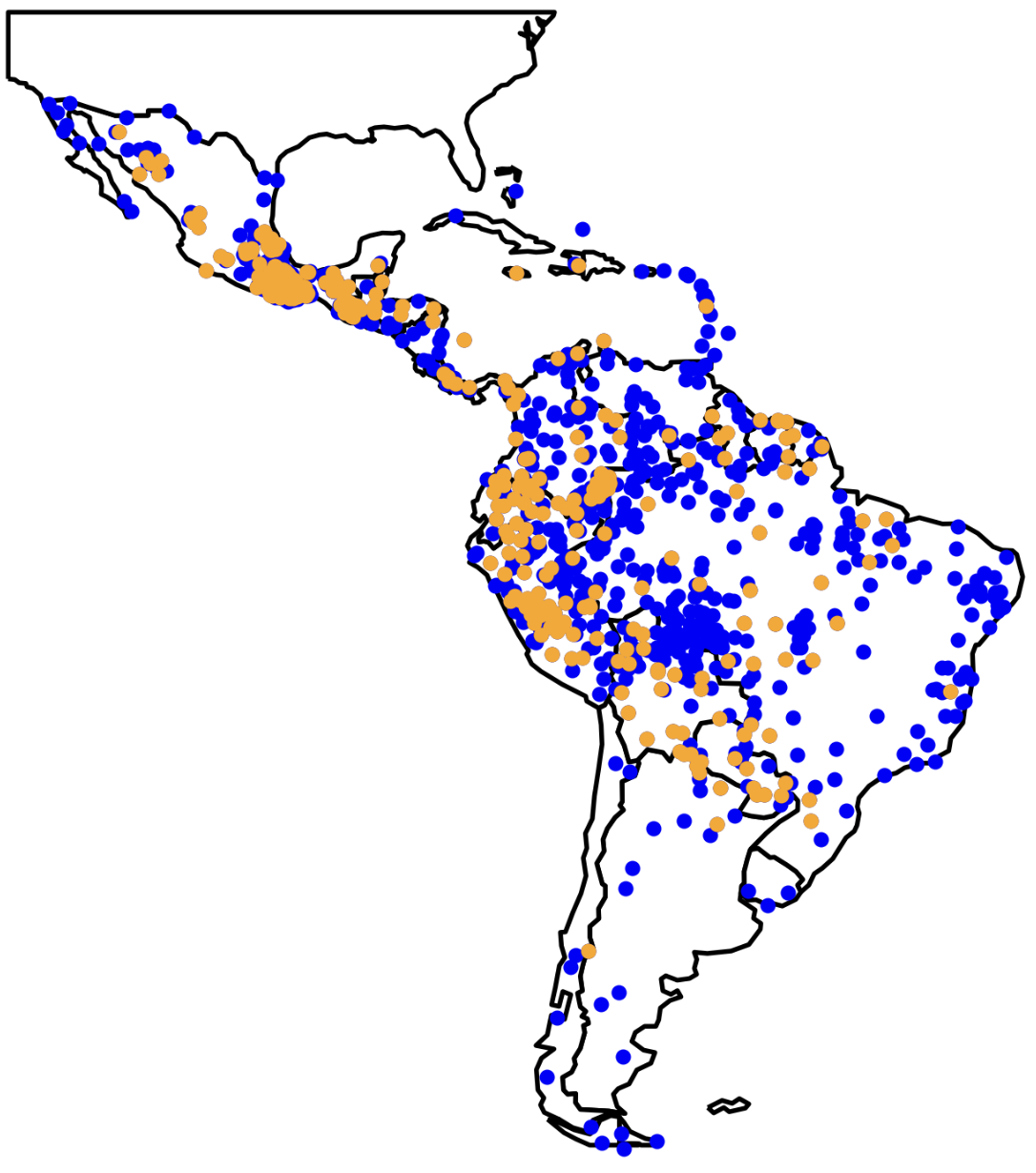}
  \caption{Approximate areal distribution of the languages in the dataset (orange) among the languages listed by Glottolog for the region (blue).}
\label{arealdistr}
\end{figure}

\paragraph{Dataset creation} The Latin American and Caribbean parallel language data used in this experiment is a subset of \citeauthor{mayer-cysouw}'s (\citeyear{mayer-cysouw}) massively parallel corpus. To identify Latin American and Caribbean varieties in the massively parallel corpus, a \verb|GeoJSON| dataset was manually created using \url{https://geojson.io/} to define the geographical region of interest. The approximate coordinates for each language variety in the dataset were taken from Glottolog and assigned to each New Testament translation based on its associated ISO 639-3 code. All varieties whose approximate coordinates were outside of the polygon defined by the \verb|GeoJSON| dataset were filtered out from the corpus. The resulting data consisted of 335 varieties, representing approximately one-third of all the languages (1,005) recorded for Latin America and the Caribbean by Glottolog.\footnote{This number excludes sign languages, as we focus on textual data.} Figure \ref{arealdistr} shows the areal distribution of the languages in our dataset among all the languages with an ISO 639-3 code from the region. 

\paragraph{Word alignment \& semantic mapping} SyMGIZA++ (\citealt{symgiza}) was used to align the English version of the New Testament to each of the translations in our dataset at the token level, achieving a one-to-one token alignment for each language (i.e. each English token corresponds to at most one token in the target language, in contrast to possible one-to-many or many-to-one alignments). The occurrences of English \textit{when} and its parallels in all Latin American and Caribbean languages in the dataset were then extracted. The quality of the automatic alignment was evaluated based on a sample of 300 \textit{when}-clauses manually aligned to the Huichol translation, against which automatic alignment achieved a precision of 0.66, recall of 1, and F1-score of 0.79.\footnote{To calculate precision and recall, the presence of an aligned word in the target language was considered a `positive', whereas the lack of an alignment (`NULL'-alignment) was considered a `negative'. For an alignment to be considered a `true negative', English \textit{when} needed to have a NULL-alignment in Huichol in cases where Huichol does not use a subjunction to render the \textsc{when}-clause, but expresses temporal subordination morphologically. Conversely, `false negatives' corresponded to any NULL-alignment which should have been aligned to a `when' word in Huichol. `False positives', then, were considered cases in which \textit{when} was aligned to a token in Huichol, despite the language using a morphological subordination strategy (i.e. switch-reference) or an independent clause, rather than a \textit{quepaucua}-(`when')-clause. Finally, `true positives' corresponded to all `when' instances correctly aligned to a `when' word in Huichol.}

\begin{figure}[!h]
  \centering
  \includegraphics[scale=0.50]{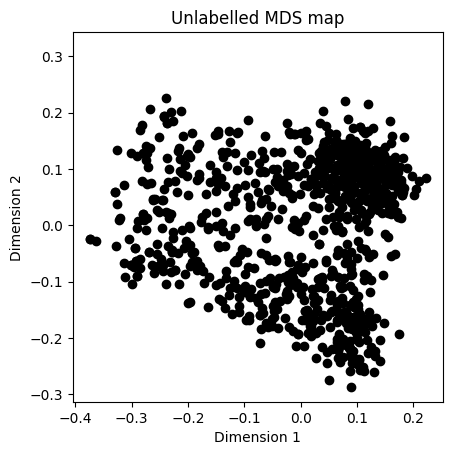}
  \caption{Unlabelled semantic map of \textsc{when}.}
\label{plainmds}
\end{figure}

Each instance of \textit{when} and its parallel in every target language was treated as one usage point for \textsc{when}. Hamming distance was applied as a measure of dissimilarity between pairs of usage points, by counting the number of languages using two different words, as opposed to the same word, for the two usage points in each pair. Multidimensional scaling (MDS) was then used to reduce the resulting Hamming-distance matrix to two dimensions, which were then treated as coordinates to plot the semantic map of \textsc{when} as shown in Figure \ref{plainmds}. Each dot in the semantic map represents a context for \textsc{when} (i.e., a New Testament verse), and the farther apart two dots are, the more different their semantics is assumed to be, and the more likely they are to be encoded by different linguistic means cross-linguistically.

\paragraph{Benchmarking} As a form of evaluation of the methods and results, this experiment leveraged detailed typological and grammatical descriptions of the morphological system of one particular Latin American language, Huichol (or Wixárika). Huichol is among the several Latin American languages that show a clear division of labor between lexified and non-lexified \textsc{when}-clauses (\citealt{anon2}). In particular, Huichol uses \textit{switch-reference} marking, a morphological system for tracking referents in an ongoing discourse (\citealt[538]{roberts_2017}). In a `canonical' switch-reference system (cf. \citealt[ix]{haimanmunro}), a clause is marked to signal whether its subject is co-referential or not with the subject of another, usually adjacent, clause, even though switch-reference has now long been shown to serve a much broader purpose than merely signaling referential (non-)identity (cf. \citealt{stirling_1993,mckenzie2012,mckenzie2015a,mckenzie2015b,keine}). With subject co-reference, a same-subject marker is used (\textsc{ss}), else a different-subject marker is employed (\textsc{ds}). Switch-reference is overwhelmingly present in languages that allow and use \textit{clause chaining}, which is the possibility of asyndetically stacking up several \textit{deranked} verb forms (\citealt{Stassen-1985, Croft-2002, Cristofaro-1998, cristofarosubordination}), that is, lacking marking of one or more tense, aspect, or mood distinctions compared to independent clauses in the same language, to signal their status as `medial' clauses or `converbs'. Switch-reference marking is well-known to serve that purpose particularly commonly among South American languages (cf. \citealt{vangijn2016srsouthame}). In other words, by capturing switch-reference markers, we also capture the morphological means (i.e. the $n$-grams, or most common morphemes) that signal subordination, in our case, specifically, temporal clauses. (\ref{canonsr}) is a Huichol example of canonical switch-reference from our dataset, where switch-reference markers are used on the dependent verb to signal its subordinate status, where the English version has a \textit{when}-clause in both cases.\footnote{In the Huichol examples, the spelling of the Bible translation in \citeauthor{mayer-cysouw}'s (\citeyear{mayer-cysouw}) corpus was kept. Note, however, that this is not the most common orthography found in most studies on Huichol today.}

\begin{example}
Huichol/Wixárika (Uto-Aztecan)
\begin{itemize}
\item[a.]
\gll Hesüana \textbf{me-'u'-axüa-cu} müpaü ti-ni-va-ru-ta-hüave 
to.him \textsc{3.pl.sbj-vis-}arrive\textsc{.pl-ds} thus \textsc{distr-narr-3.pl.nsbj-pl-sg-}say
\glt `When they came to him he said to them' (Acts 20:18)
\glend
\item[b.]
\gll Hesüana \textbf{me-'u'-axüa-ca} müme, müpaü me-te-ni-ta-hüave
to.him \textsc{3.pl.sbj-vis-}arrive\textsc{.pl-ss} men thus \textsc{3.pl.sbj-distr-narr-sg-}say
\glt ‘When the men had come to him they said’ (Luke 7:20)
\glend
\label{canonsr}
\end{itemize}
\end{example}

\noindent Huichol additionally has a lexified `when' subordinator (\textit{quepaucua}), in which case switch-reference marking is absent, as in (\ref{quepaucua}).

\begin{example}
\gll Mericüsü \textbf{quepaucua} yemuri-sie \textbf{m-a-ca-ne}, teüteri yumüiretü me-ca-n-i-veiya-caitüni
then when mountain-\textsc{loc} \textsc{as2}-\textsc{pro}-down-arrive.\textsc{pfv}, people many \textsc{3.pl.sbj}-\textsc{narr}-\textsc{narr}-\textsc{3.sg.obj}-follow-\textsc{ipfv}
\glt `When he came down from the mountain, great crowds followed him' (Matthew 8:1)
\glend
\label{quepaucua}
\end{example}

\begin{figure}[H]
  \centering
  \includegraphics[scale=0.50]{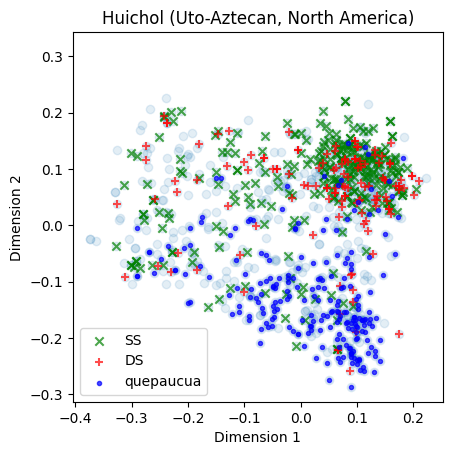}
  \caption{Probabilistic semantic map of \textsc{when}, showing the location of lexified subordinators and switch-reference markers in Huichol after direct annotation (used as benchmark).}
\label{huicholnokrigold}
\end{figure}

\noindent The concurrent presence of both a lexified connector and easily isolable morphemes for morphological subordination makes the language an ideal initial benchmark for experimenting with automatically detecting morphological \textit{and} lexified markers of temporal subordination in the parallel corpus. As a form of evaluation for the character $n$-gram search system described below, the Huichol translation of the New Testament was enriched with annotation for different switch-reference markers. The markers were identified by using existing descriptions of Huichol switch-reference (i.e. \citealt{comriehuichol, grammaticalrelhuichol, huicholthesis}). The language has easily isolable switch-reference morphemes, namely \textit{-ku} and -\textit{ka} (spelled as \textit{-cu} and \textit{ca} in our dataset), for `different-subject' and `same-subject' marker, so the placeholders \textsc{ds} and \textsc{ss} were inserted before any word in the Huichol text ending with the respective forms, thus allowing the alignment model to capture the placeholders as dummy subordinators. Based on the annotated dataset, the location of \textsc{ss} and \textsc{ds} markers in the semantic map (Figure \ref{huicholnokrigold}) can be compared with the location of morphological markers identified automatically via character $n$-gram search (Figure \ref{huichol} in Section \ref{sec:maps}).

\paragraph{N-gram search} Character $n$-grams were leveraged to identify potential morphological markers that are highly correlated with English \textit{when}-clauses in our dataset, \textit{in addition} to lexified means. As mentioned in the Introduction, the identification of potentially meaningful $n$-grams (i.e. those expressing a particular meaning of \textsc{when}) is based on the approach by \citet{superpivot}, albeit with additional steps and different $n$-gram ranges. Similarly to \citet{superpivot}, $\chi^2$ is used as a score of association between a `head pivot' (in our case always \textit{when}) and a character $n$-gram, and it is calculated based on how many times \textit{when} is aligned to a word containing that $n$-gram, how many times it is aligned to other $n$-grams and the frequency of both \textit{when} and the $n$-gram. The raw alignments by SymGIZA++ were used as a starting point to identify tokens on which the $n$-gram search should be carried out. The following steps were followed to subsequently refine the parallel dataset with potentially meaningful $n$-grams:
\begin{itemize}
    \item[1.] a bespoke list of stopwords in English was established, based on their being either extremely frequent (\textit{Jesus}, \textit{Herod}, \textit{Peter}, \textit{Paul}) or very likely to introduce noise in a study on temporal subordination because of their distributional overlap with subordinators in terms of absolute position in a sentence (\textit{and}, \textit{behold}, \textit{then}). $\chi^2$ was used to find highly associated forms and parallel forms with an associated $p$-value of $0$ were removed from the target language;
    \item[2.] associations were identified between \textit{when} and all tokens aligned to \textit{when} by SymGIZA++. Only tokens with the highest score and a $p$-value of $0$ were kept as they were and did not undergo the next steps; 
    \item[3.] using spaCy's (\citealt{spacy2}) English model \verb|en_core_web_sm|, the English source text was automatically annotated for syntactic dependency to identify the head of the token \textit{when}. This allowed for the \textit{verb} of the \textit{when}-clause to be extracted and the parallel verb in the translation to be identified. This choice was informed by the observation that languages marking subordination on the verb itself (i.e. non-lexified \textsc{when}-clauses) are much more likely to have an empty token <NOMATCH> aligned to English \textit{when} rather than the verb itself, so that the latter must be included in the search for meaningful character $n$-grams associated with \textit{when};
    \item[4.] associations were identified between \textit{when} and $n$-grams of any size between 2 and 9 for all remaining tokens aligned to either \textit{when} or its head verb;
    \item[5.] the top-scoring 200 $n$-grams (by $\chi^2$) were then sorted by the number of times \textit{when} was found to cooccur with the $n$-gram. The top-scoring 20 $n$-grams among the latter were then extracted as potentially meaningful $n$-grams;
    \item[6.] the 20 extracted $n$-grams were clustered to attempt capturing groups of $n$-grams that are likely to be allomorphs of the same morpheme. Clustering was done using DBSCAN after converting the list of $n$-grams to a matrix of TF-IDF features. DBSCAN was selected after comparison with several other clustering algorithms (i.e. K-Means, K-Means++, Agglomerative Clustering, and Gaussian Mixture Modelling).
    \item[7.] Each cluster of $n$-grams was assigned the placeholder label \verb|ngram_1|...\verb|ngram_N|, where \verb|N| is the number of potentially meaningful $n$-gram clusters found for any given language.
\end{itemize}

\paragraph{Geostatistical interpolation} Ordinary Kriging was then used to interpolate the linguistic items (i.e. the parallel token, if any, to \textit{when}, or the $n$-gram placeholder label) used in each data point by each language in the dataset, to look for semantically relevant cross-linguistic dimensions. The Kriging model was implemented using the PyKrige library (\citealt{pykrige}), with a Gaussian variogram model, a single averaging bin for the variogram (\verb|nlag|), and \verb|coordinates_type| set to \verb|geographic|. The optimal \verb|range|, \verb|sill|, and \verb|nugget| values for the Kriging models were set through a trial-and-error calibration process. Different combinations of these parameters were tested, and the ones used to produce the maps presented in Section \ref{sec:maps} were chosen based on the interpretability of the resulting contour maps, with particular attention to the map for Huichol, thanks to the additional automatic annotation performed on the language using external knowledge bases. 
The contour levels generated through Kriging were normalized between 0 and 1 to facilitate the interpretation of the relative intensity of a linguistic means in the semantic space so that the closer the contour level to 1, the more intense the concentration of the respective means in the area. In the maps in Section \ref{sec:maps}, contours are plotted at all levels between 0.8 and 1. 

The advantage of employing a geostatistical approach, such as Ordinary Kriging, for mapping language patterns is its ability to account for spatial autocorrelation (cf. \citealt{getisautocorr}), which facilitates the nuanced weighting of variables based on their prevalence and intensity across geographical space. While one linguistic means might be more widespread in terms of raw occurrence count in a given region of the semantic map, Kriging allows us to discern the spatial intensity of competing means. This, in turn, can clarify whether other means, despite being less prevalent overall, are more concentrated in that area and therefore more directly representative of the meaning associated with the respective space in the semantic map. 

In the Kriging maps in Section \ref{sec:maps}, the placeholders for the $n$-grams are used instead of the actual list of $n$-grams.\footnote{The reader can find which $n$-grams each group contains for any given language in the \href{https://doi.org/10.6084/m9.figshare.25431814.v1}{associated repository}.}

\section{Results}\label{sec:maps}
\begin{figure}[H]
  \centering
  \includegraphics[scale=0.20]{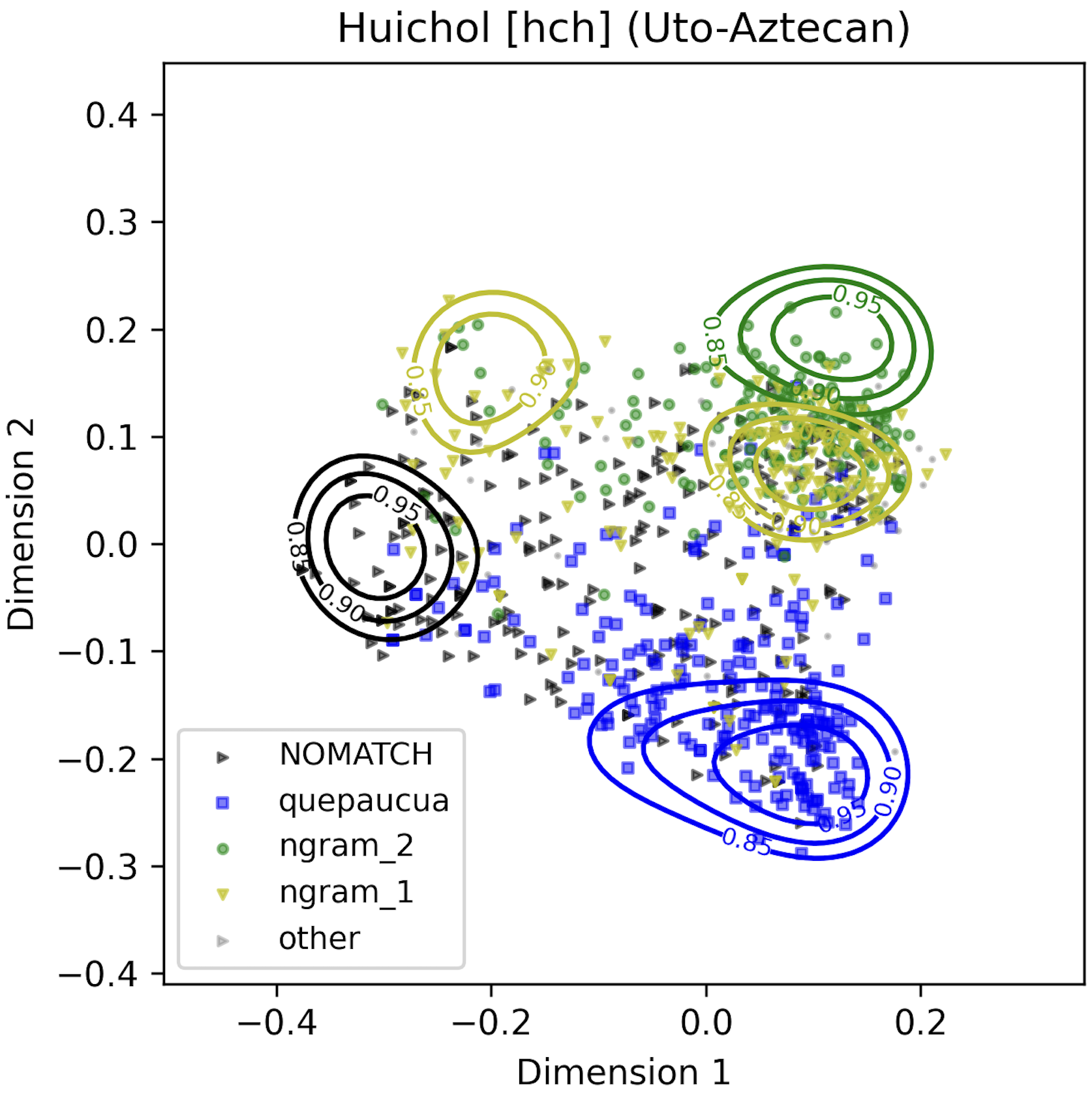}
  \caption{Kriging map of \textsc{when} for Huichol.}
\label{huichol}
\end{figure}

\paragraph{Huichol} Figure \ref{huichol} shows the Kriging map generated from the Huichol data automatically refined with the $n$-gram search method. This can be compared with the labeled map in Figure \ref{huicholnokrigold}, which, as explained in the previous section, is instead based on the Huichol data directly annotated with switch-reference markers as presented in typological descriptions of the language.

\noindent Kriging detected relatively clearly separate areas (i.e. contexts or usage points) for lexified means (\textit{quepaucua}), clustering at the bottom right of the map, and non-lexified means, corresponding to \textit{ngram\textunderscore1} and \textit{ngram\textunderscore2} in the map and clustering at the top of the map. \textit{NOMATCH} indicates the absence of a parallel to English \textit{when}, which suggests either a misalignment or the usage of a non-subordinate construction (e.g. an independent clause or a prepositional phrase, e.g. `during dinner'). It is clear that the two automatically identified groups of $n$-grams, \textit{ngram\textunderscore1} and \textit{ngram\textunderscore2}, in the Huichol map correspond to \textsc{ds} and \textsc{ss} markers respectively. The \textit{ngram\textunderscore1} group includes \textit{u}, \textit{su}, \textit{usu}, \textit{cusu}, \textit{icusu}, \textit{ricusu}, \textit{ericusu}, whereas \textit{ngram\textunderscore2} includes \textit{ca}, \textit{aca}, \textit{eca}, \textit{ieca}, \textit{yaca}, \textit{iyaca}, \textit{xeiyaca}, \textit{eiyaca}, \textit{nieca}, which match the known switch-reference markers \textit{-ku} and -\textit{ka} (spelled as \textit{-cu} and \textit{ca} in our dataset) for \textsc{ds} and and \textsc{ss} respectively (\citealt{comriehuichol}).

Based on the Huichol results, automatic word-alignment combined with the $n$-gram search method achieves a precision of 0.90, recall of 0.99, and F1-score of 0.94, calculated upon comparison with another manually annotated random sample of 300 English-Huichol \textsc{when}-clauses with added switch-reference distinctions (i.e. English \textit{when} was manually aligned to either \textit{quepaucua} `when', \textsc{ds}, or \textsc{ss}).

\begin{figure*}[ht!] 
  \centering
  \begin{subfigure}[b]{0.30\textwidth}
    \centering
    \includegraphics[width=\textwidth]{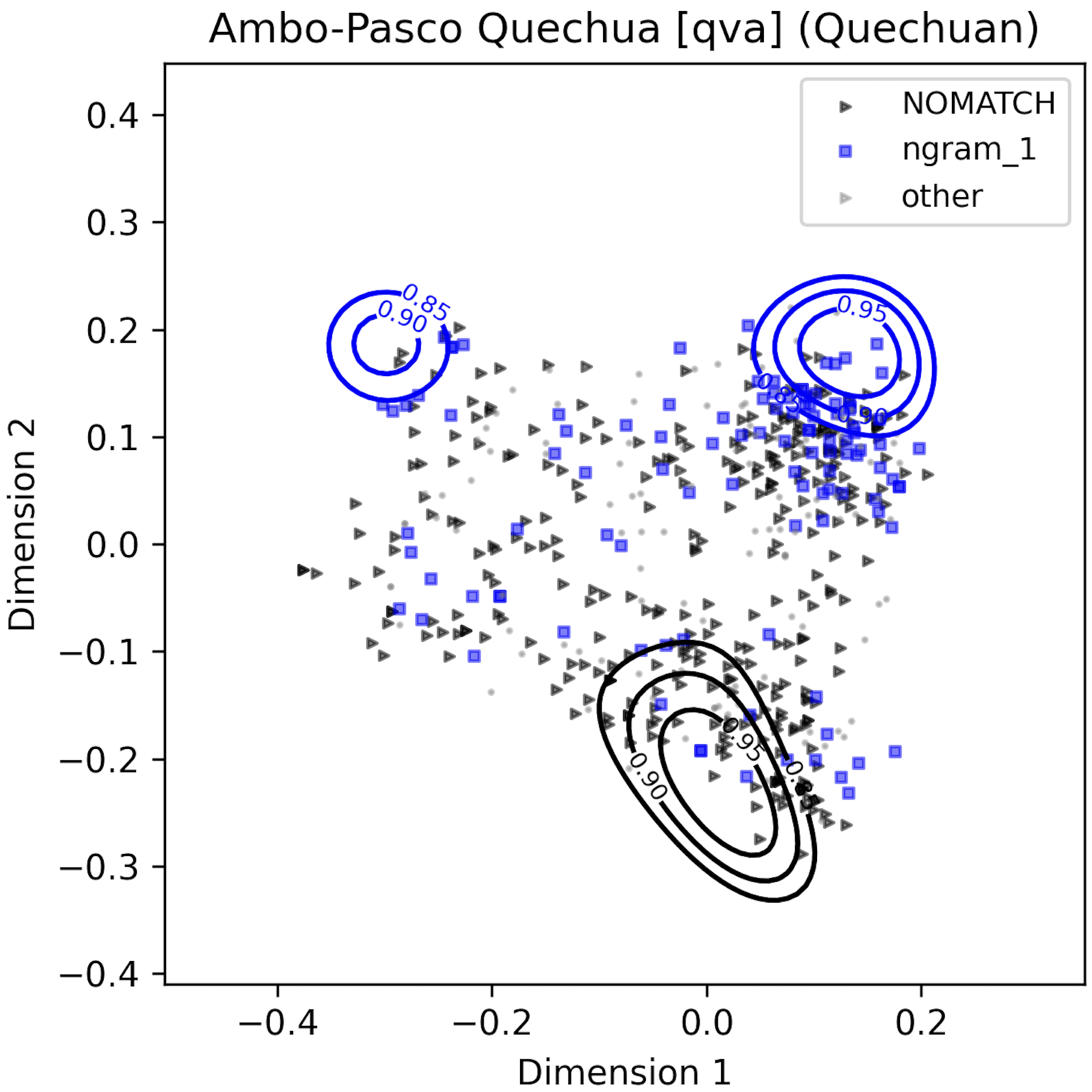}
    \caption{}
    \label{qva}
  \end{subfigure}
  \begin{subfigure}[b]{0.30\textwidth}
    \centering
    \includegraphics[width=\textwidth]{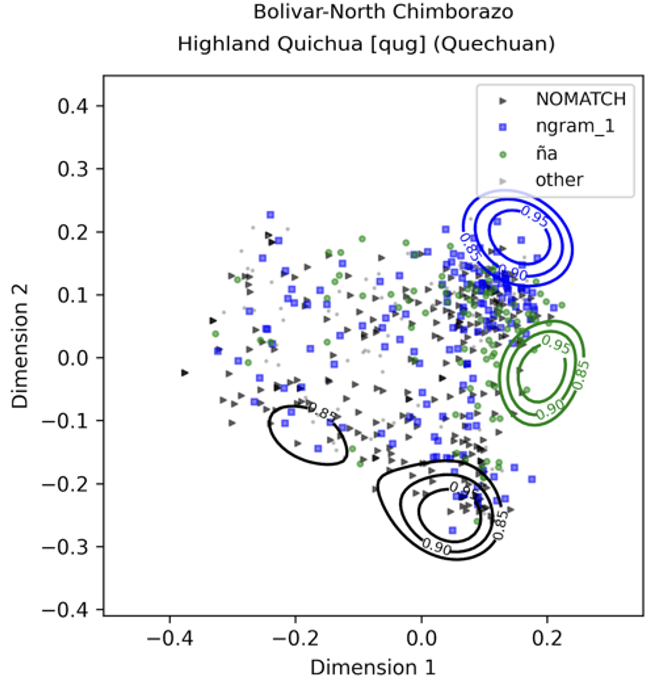}
    \caption{}
    \label{qug}
  \end{subfigure}
  \begin{subfigure}[b]{0.30\textwidth}
    \centering
    \includegraphics[width=\textwidth]{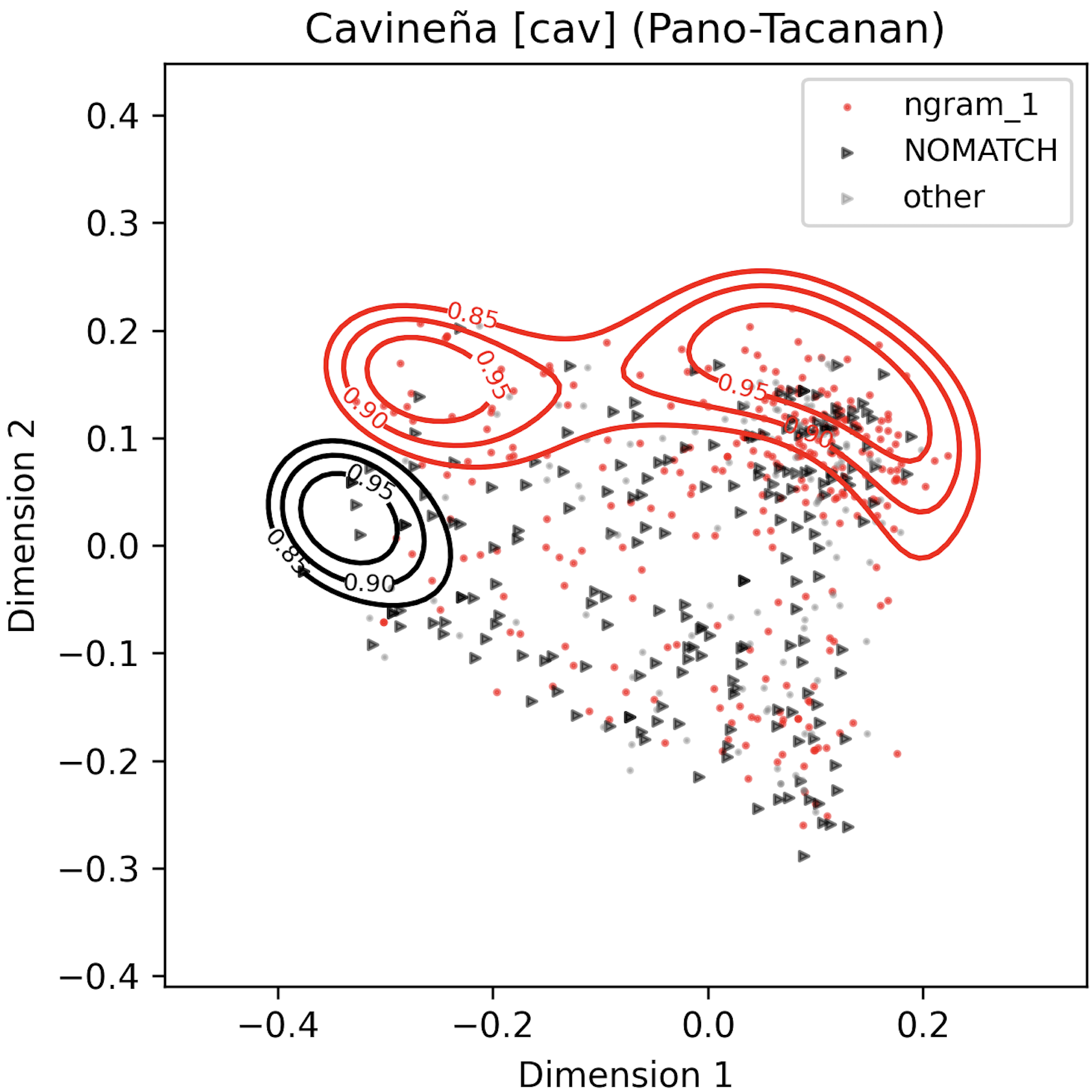}
    \caption{}
    \label{cav}
  \end{subfigure}
  \caption{Kriging maps of \textsc{when} for three Latin American languages.}
  \label{fig:threelangs}
\end{figure*}

\paragraph{Switch-reference languages} A clear validation of our method comes from the Quechuan languages in our dataset. According to \citet[168-169]{vangijn2016srsouthame}, all Quechuan languages have switch-reference marking, albeit with some differences in the markers used and their semantic scope. A closer inspection of the maps reveals that all Quechuan languages in our dataset show, in fact, a clear division of labor between the bottom and top of the map. Most commonly, the former is a \textit{NOMATCH} area, whereas the top areas are instead most clearly under the scope of switch-reference markers. This is clearly the case, for example, in Ambo-Pasco Quechua (Figure \ref{qva}), from Peru, where the \textit{ngram\textunderscore1} group at the top of the map includes \textit{r}, \textit{ar}, \textit{ur}, \textit{cur}, \textit{ycur}, \textit{aycur}, \textit{car}, all of which contain the distinctive \textit{-r} \textsc{ss} marker of some Quechuan I subgroups (cf. \citealt[168]{vangijn2016srsouthame}).

Another example is the map for Bolivar-North Chimborazo Highland Quichua (Figure \ref{qug}), from Ecuador. In this case, an \textit{ngram\textunderscore1} Kriging area was detected alongside a potentially lexified subordinator \textit{ña}. The $n$-gram group includes \textit{aca}, \textit{paca}, \textit{hpaca}, \textit{shpaca}, \textit{ushpaca}, \textit{ashpaca}, where the Quechuan II \textsc{ss} marker, /\textipa{S}/, spelled \textit{sh}, can be discerned (\citealt[171]{vangijn2016srsouthame}).\footnote{The -\textit{ca} ending is, in all likelihood, a personal ending that is particularly frequent in the source text.}

A similar split, where the top area of the map is dominated by a $n$-gram group, is also found outside of Quechuan. This is the case, for instance, of Cavineña (Figure \ref{cav}), a Pano-Tacanan language of the Amazonian plains of northern Bolivia, where \textit{ngram\textunderscore1} includes \textit{u}, \textit{su}, \textit{tsu}, \textit{atsu}, \textit{aatsu}, \textit{catsu}, \textit{baatsu}, \textit{acatsu}, \textit{bacatsu}, \textit{itsu}, where the \textsc{ss} marker \textit{-tsu} (cf. \citealt{guillaume2008,guillaume2011}) can be seen.

\begin{figure*}[ht!] 
 \centering
  \begin{subfigure}[b]{0.30\textwidth}
  \centering
    \includegraphics[width=\textwidth]{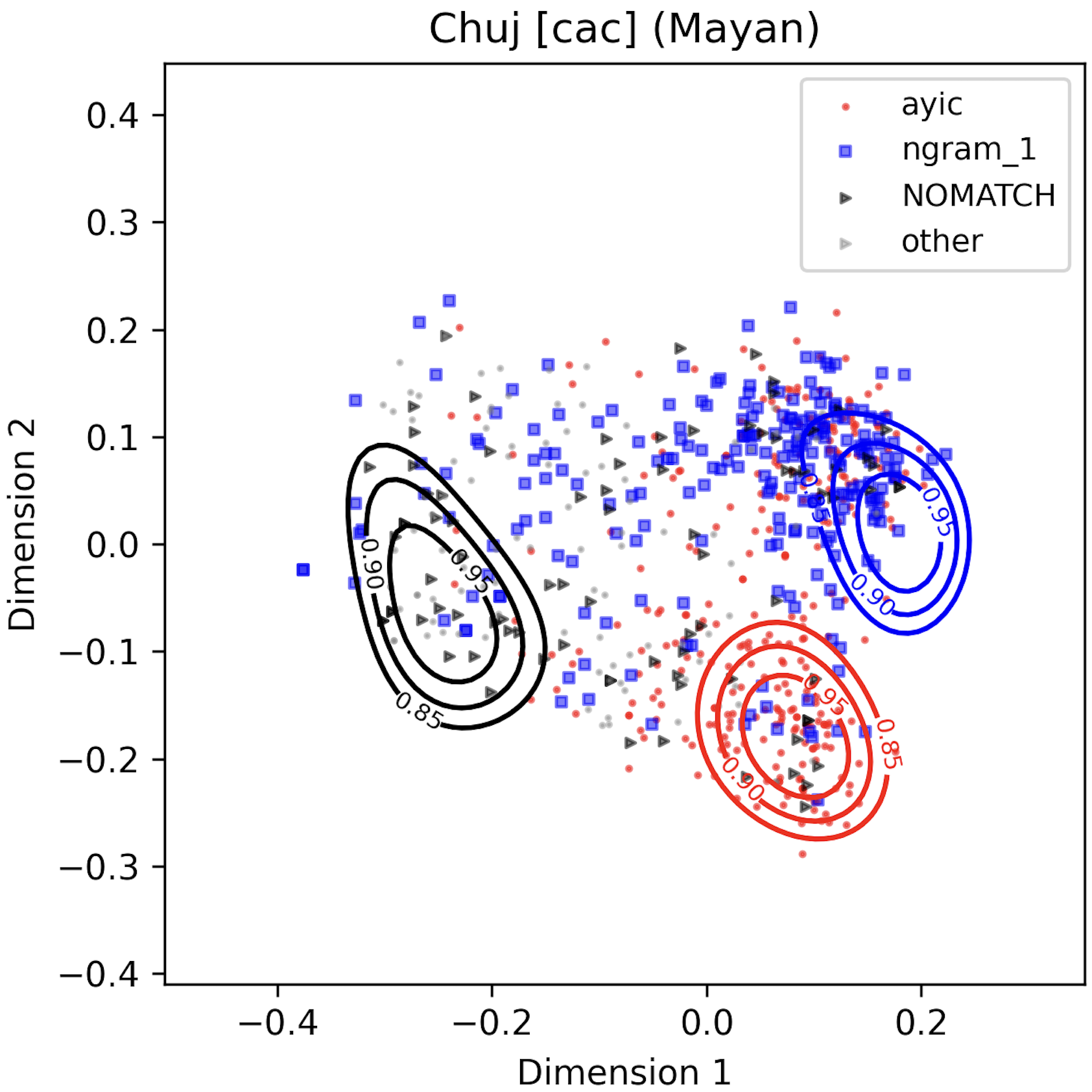}
    \caption{}
    \label{cac}
  \end{subfigure}
  \centering
  \begin{subfigure}[b]{0.30\textwidth}
  \centering
    \includegraphics[width=\textwidth]{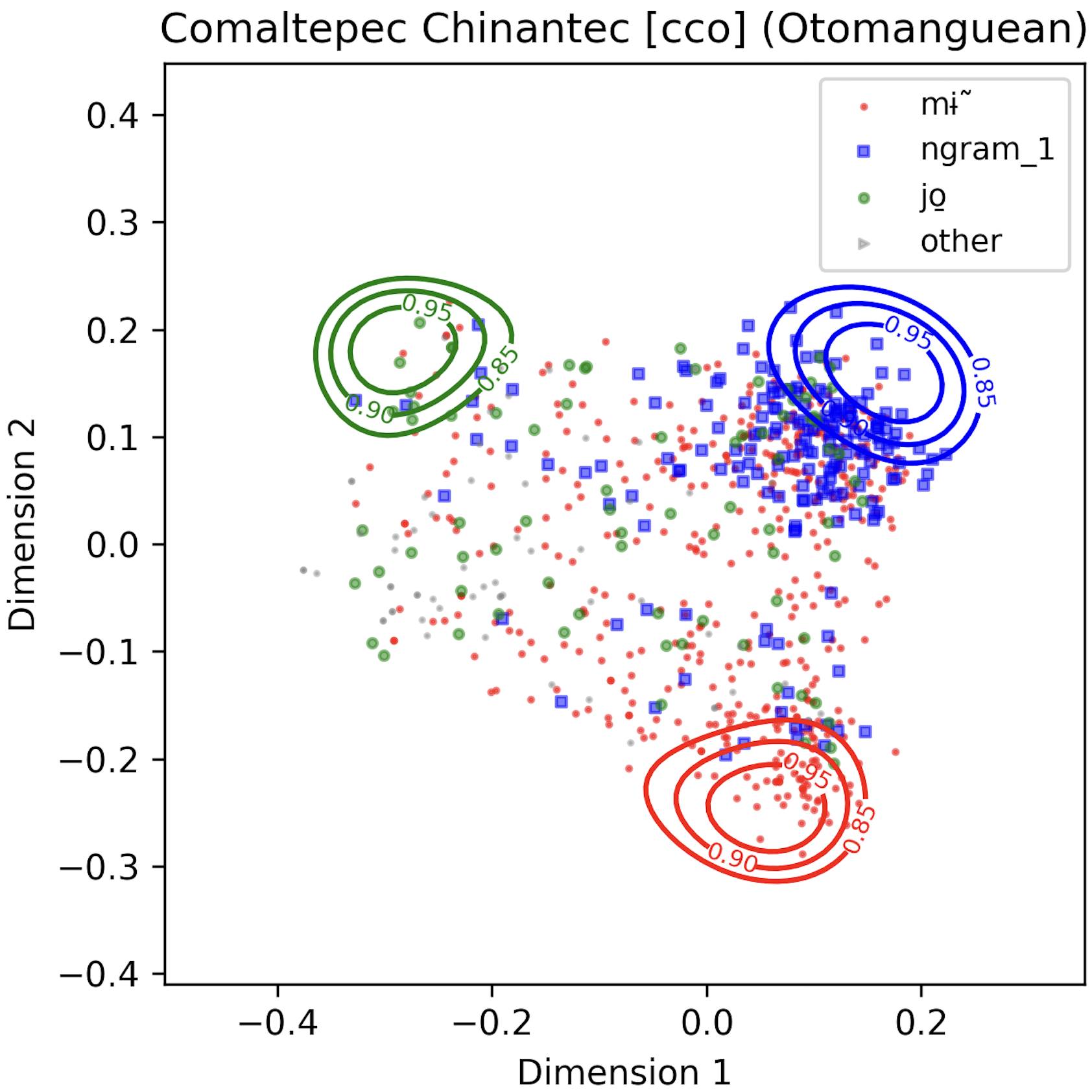}
    \caption{}
    \label{cco}
  \end{subfigure}
  \centering
  \begin{subfigure}[b]{0.30\textwidth}
  \centering
    \includegraphics[width=\textwidth]{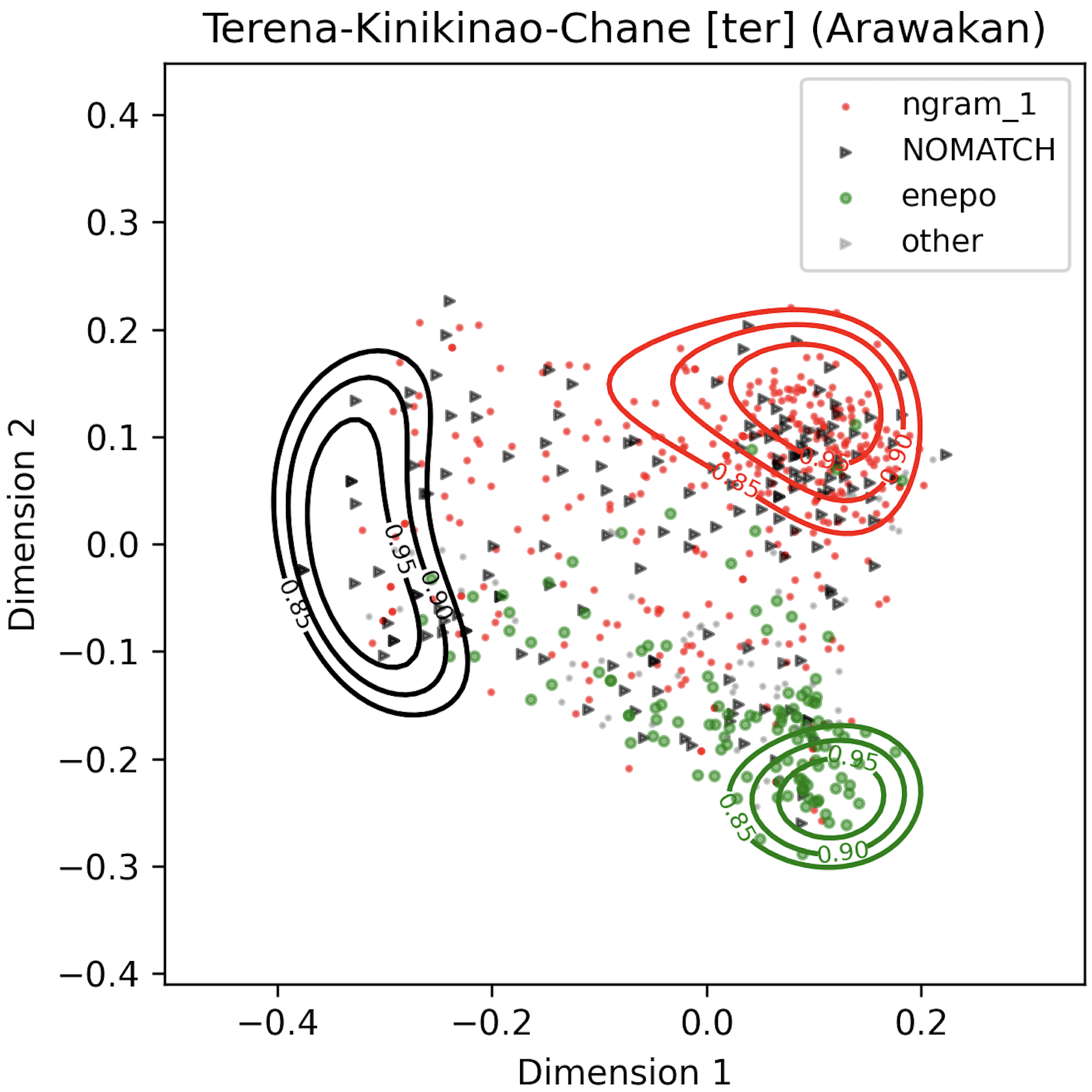}
    \caption{}
    \label{ter}
  \end{subfigure}
  \centering
  \begin{subfigure}[b]{0.30\textwidth}
    \includegraphics[width=\textwidth]{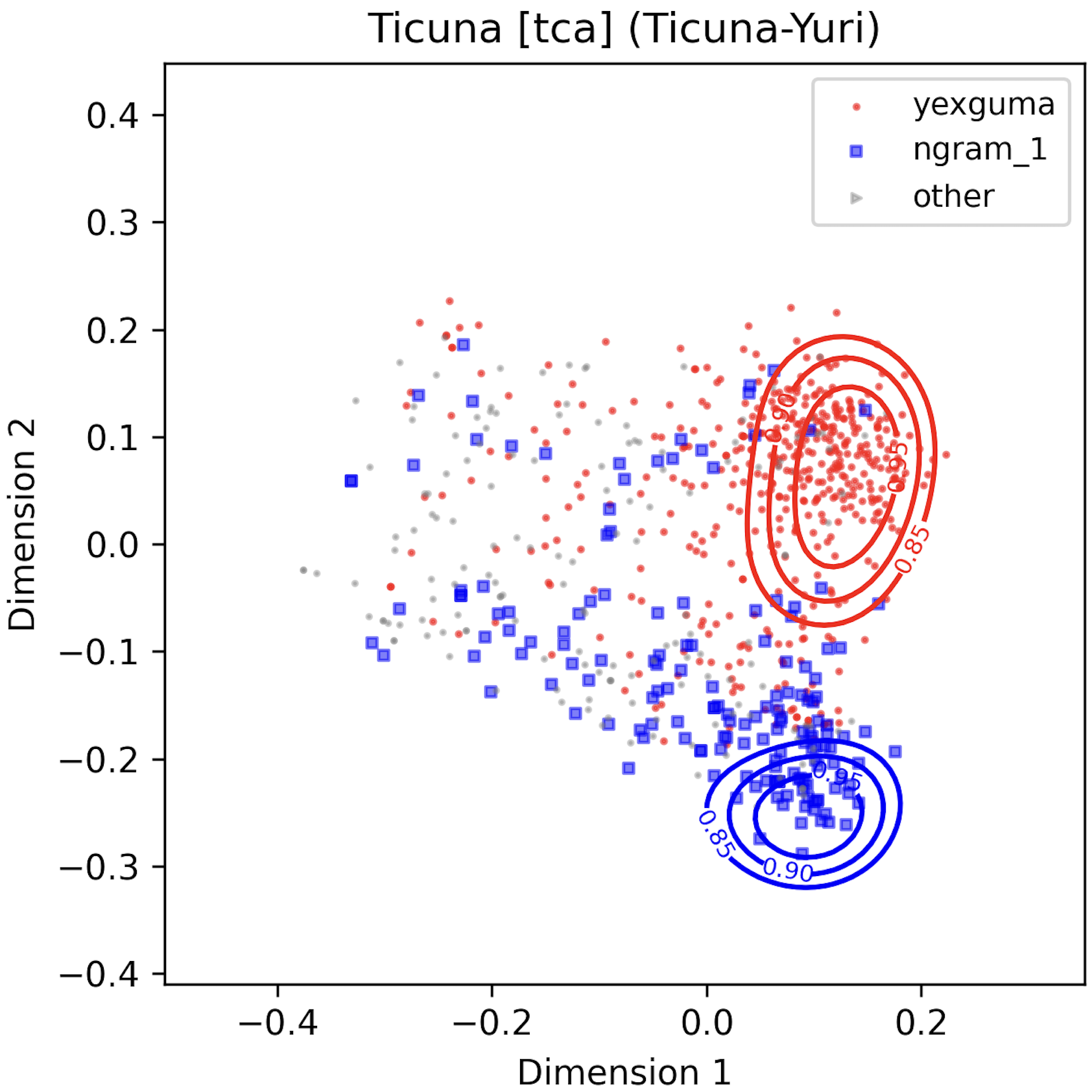}
    \caption{}
    \label{tca}
  \end{subfigure}
  \centering
  \begin{subfigure}[b]{0.32\textwidth}
    \includegraphics[width=\textwidth]{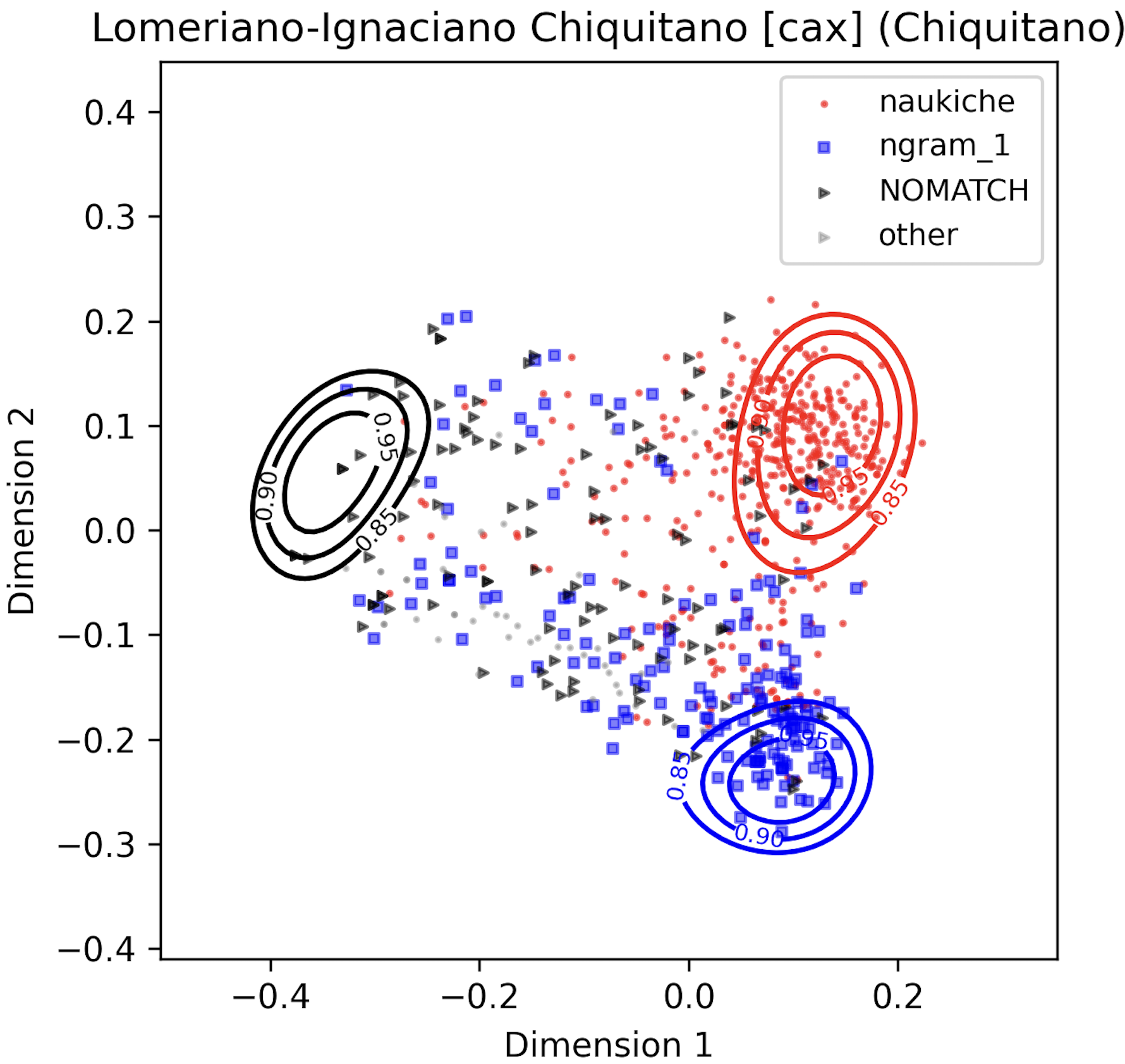}
    \caption{}
    \label{cax}
  \end{subfigure}
  \centering
  \begin{subfigure}[b]{0.30\textwidth}
    \includegraphics[width=\textwidth]{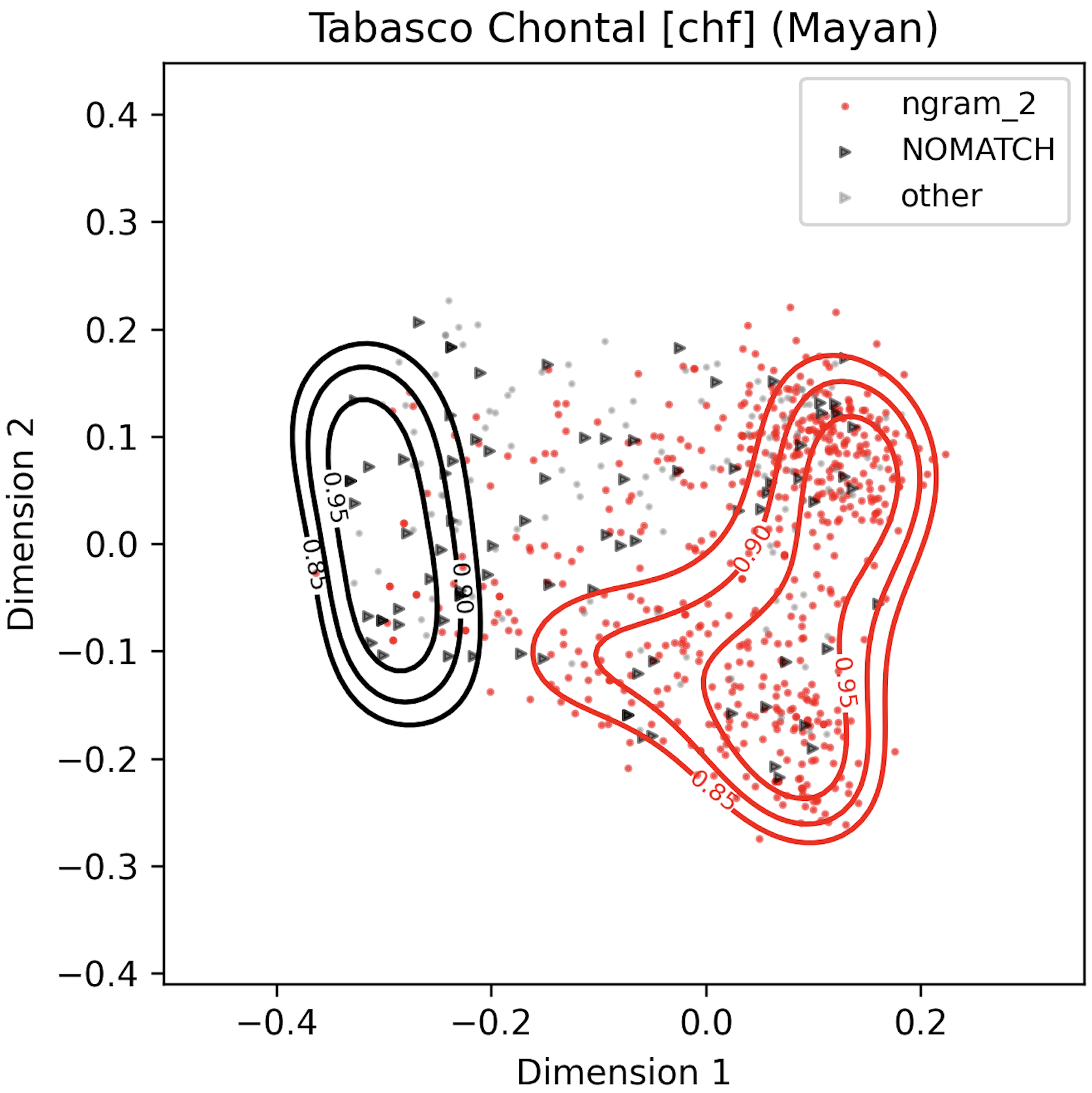}
    \caption{}
    \label{chf}
  \end{subfigure}
  \centering
  \begin{subfigure}[b]{0.30\textwidth}
    \includegraphics[width=\textwidth]{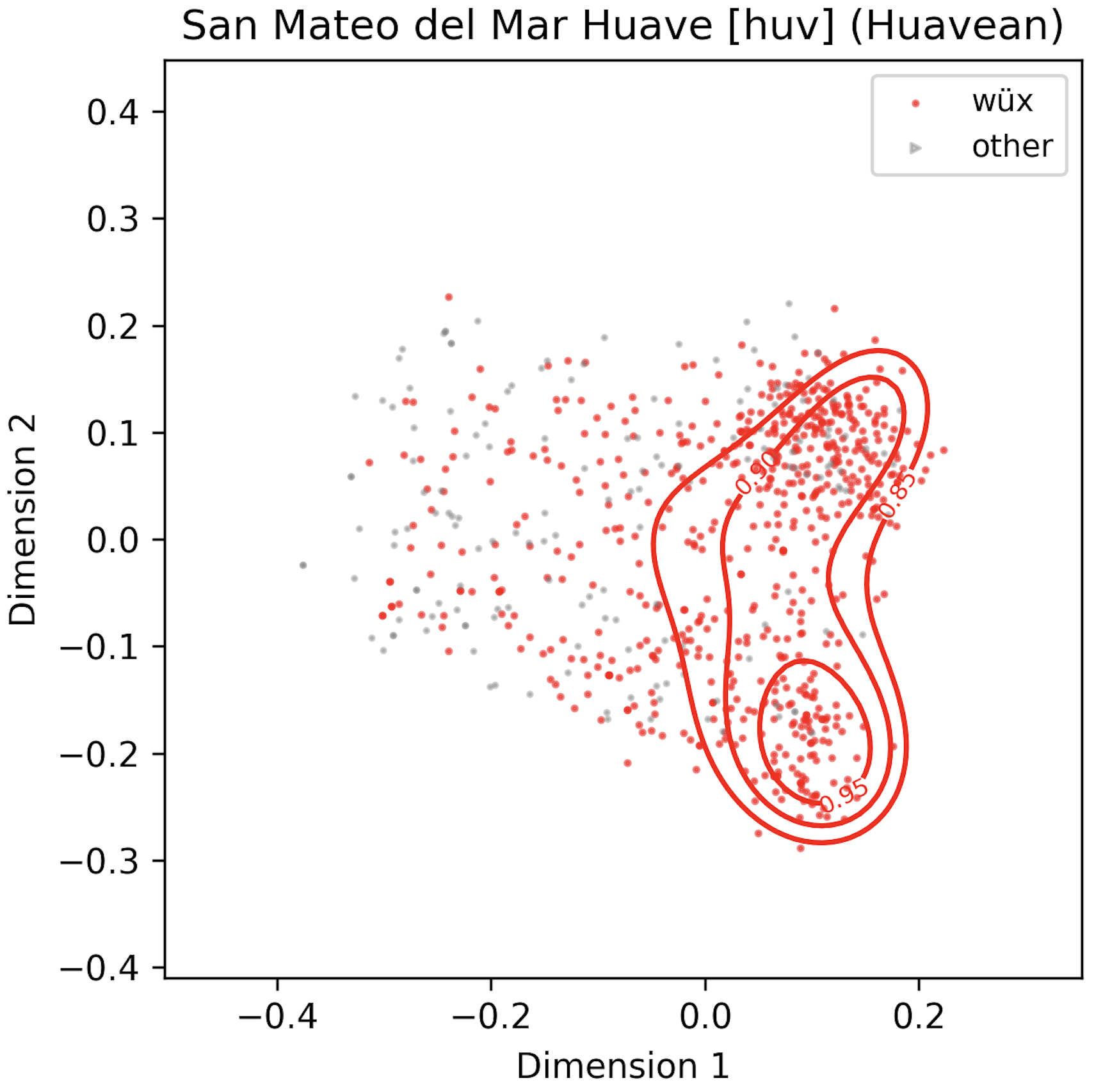}
    \caption{}
    \label{huv}
  \end{subfigure}
  \centering
  \begin{subfigure}[b]{0.30\textwidth}
    \includegraphics[width=\textwidth]{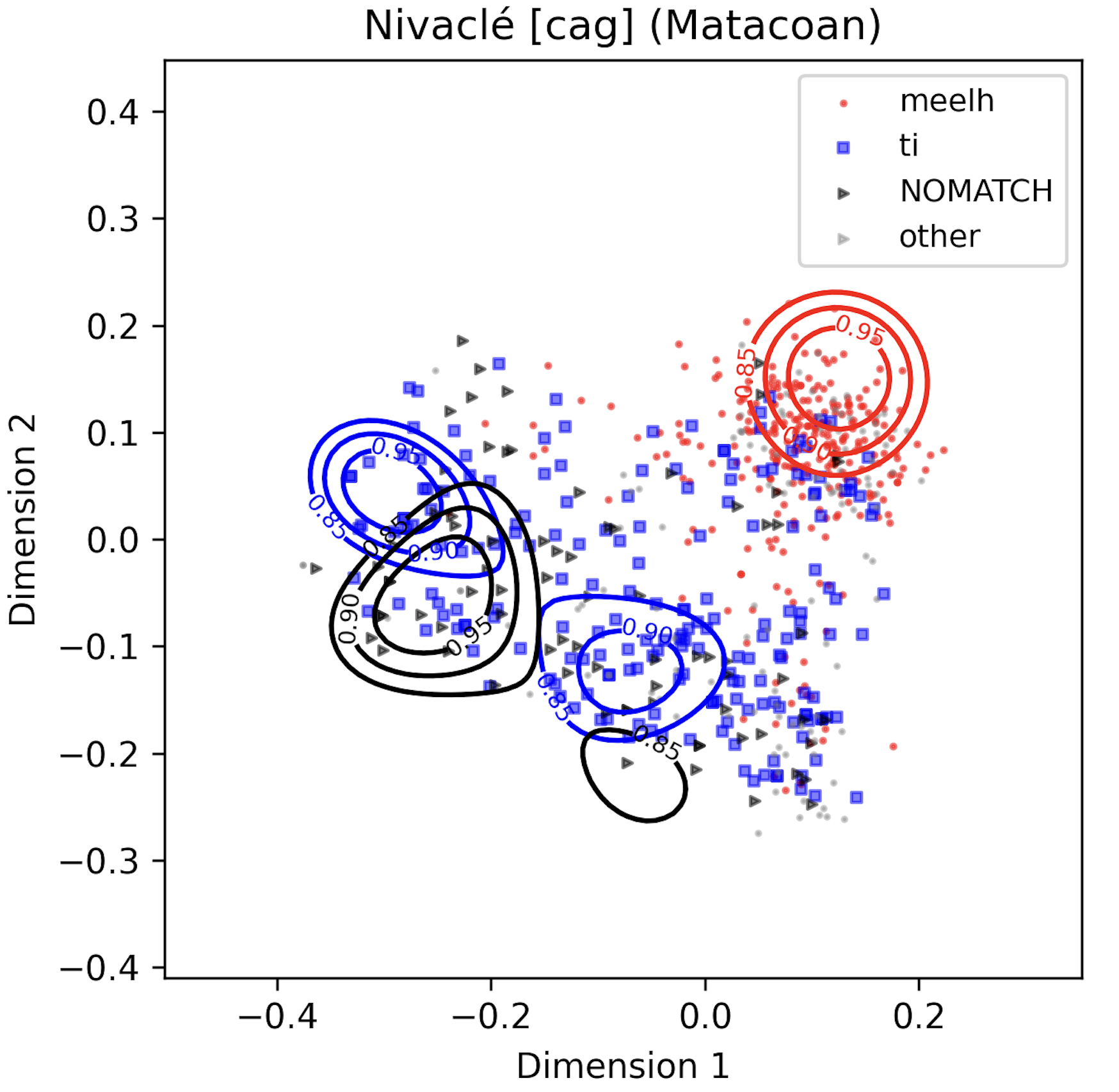}
    \caption{}
    \label{cag}
  \end{subfigure}
  \centering
  \begin{subfigure}[b]{0.30\textwidth}
  \centering
    \includegraphics[width=\textwidth]{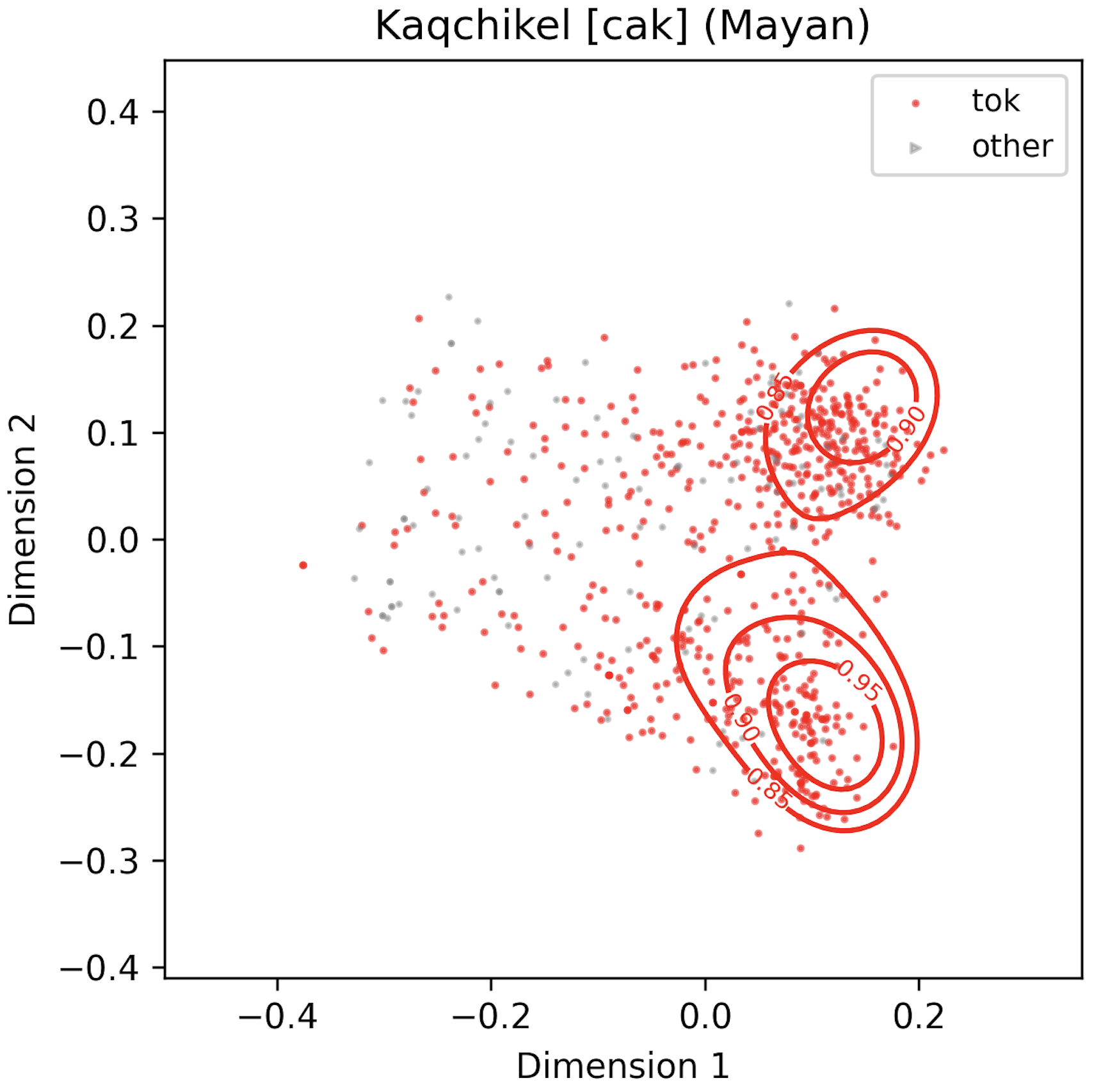}
    \caption{}
    \label{cak}
  \end{subfigure}
  \centering
  \begin{subfigure}[b]{0.30\textwidth}
    \includegraphics[width=\textwidth]{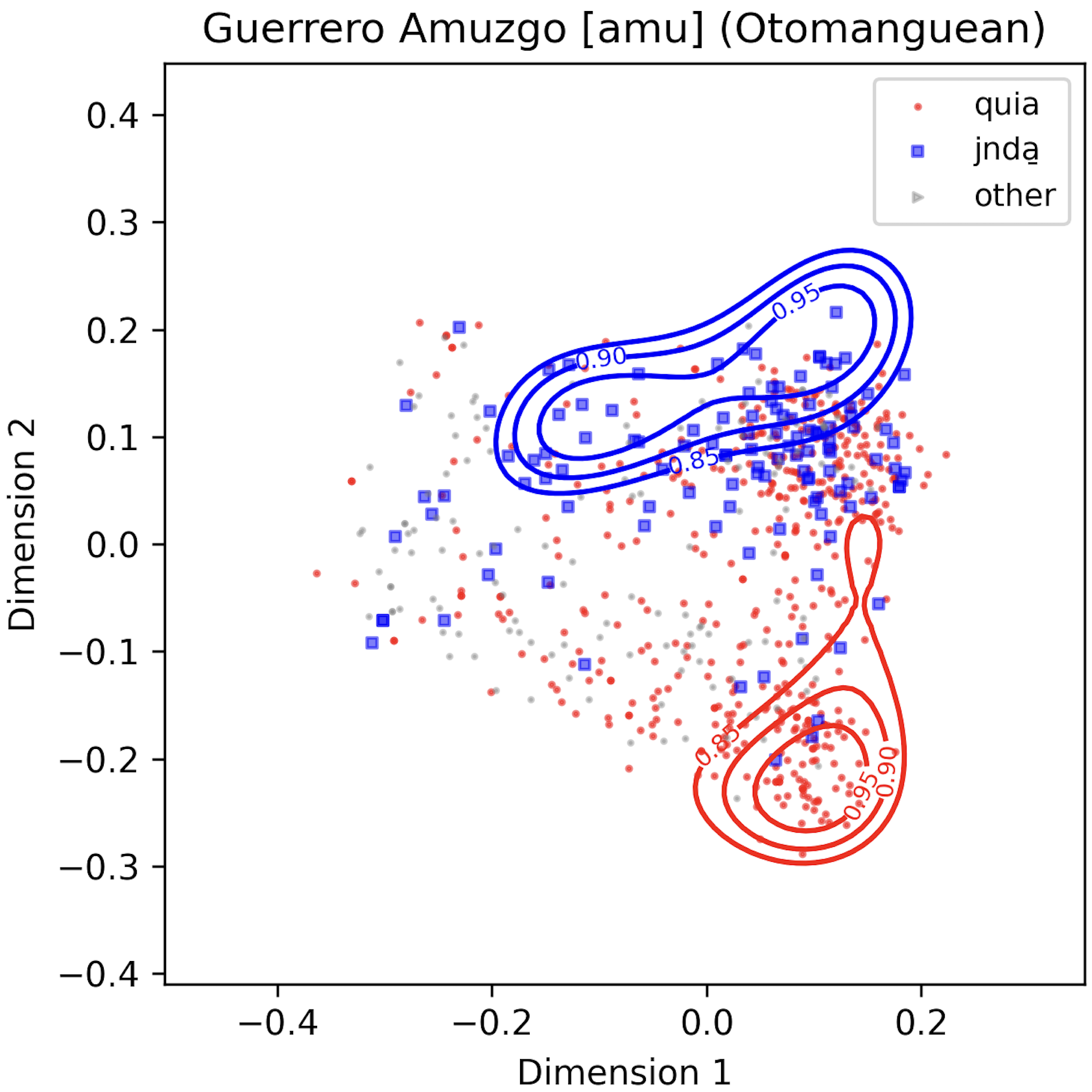}
    \caption{}
    \label{amu}
  \end{subfigure}
  \centering
  \begin{subfigure}[b]{0.30\textwidth}
    \includegraphics[width=\textwidth]{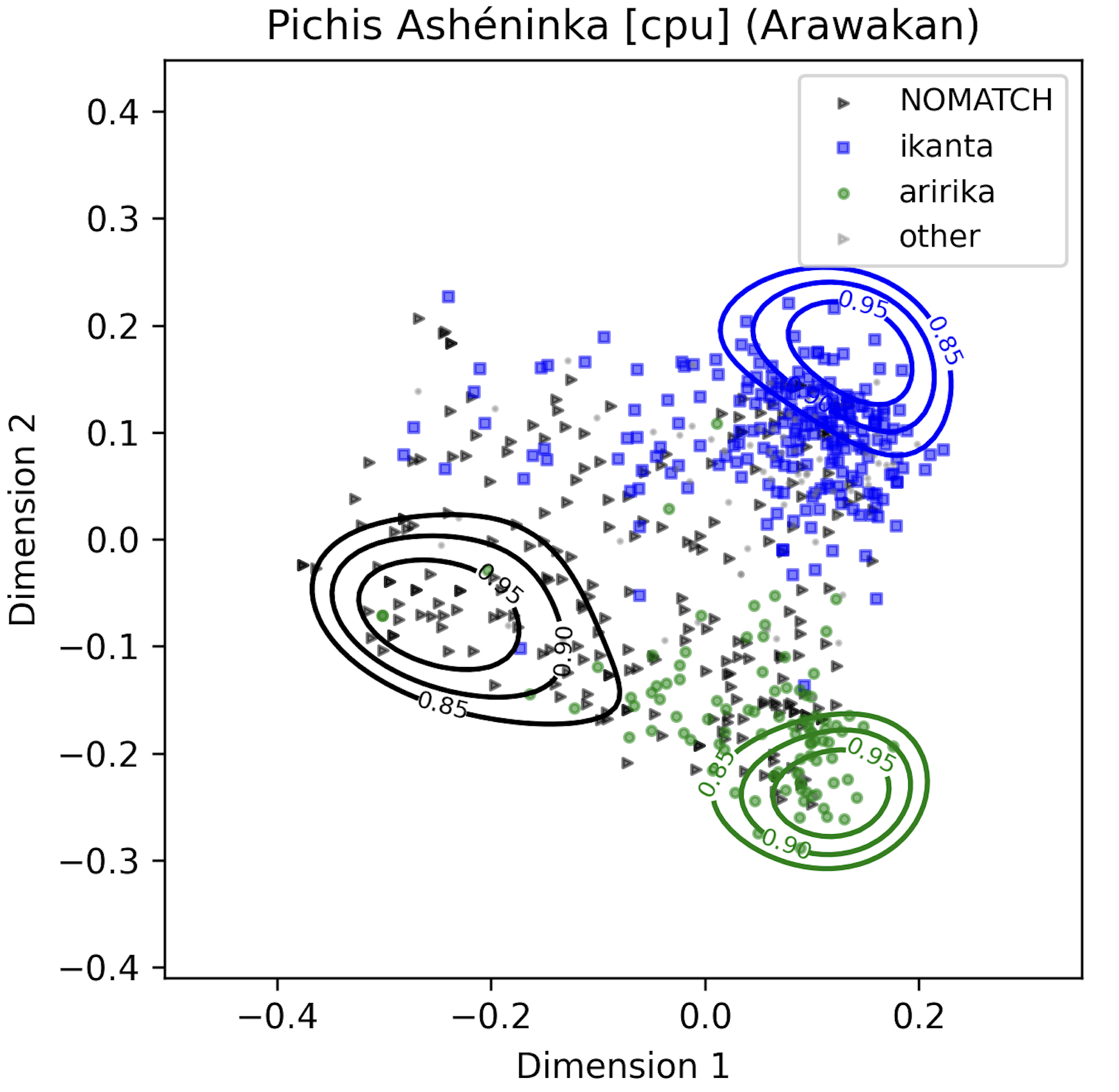}
    \caption{}
    \label{cpu}
  \end{subfigure}
    \centering
  \begin{subfigure}[b]{0.30\textwidth}
    \includegraphics[width=\textwidth]{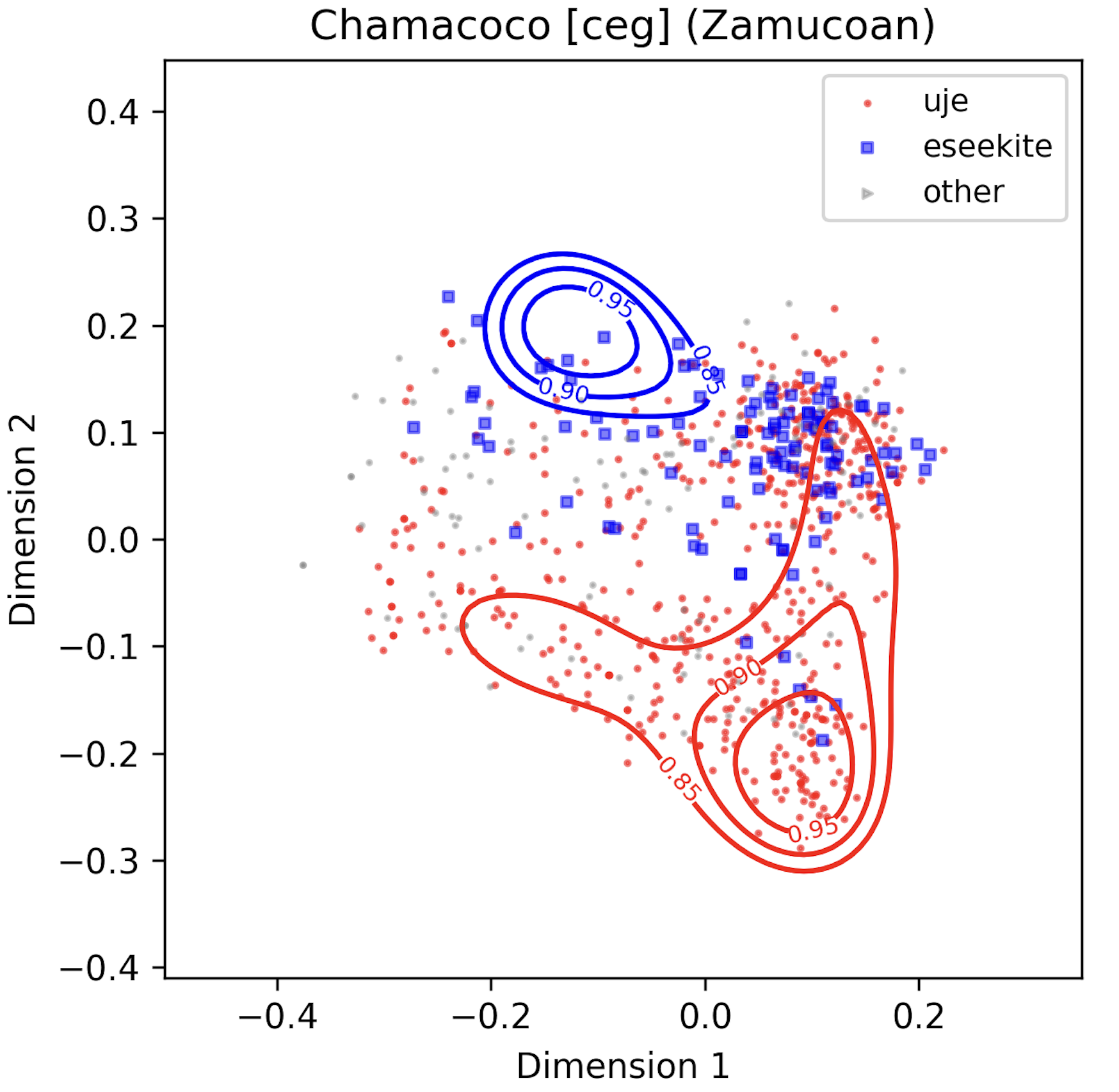}
    \caption{}
    \label{ceg}
  \end{subfigure}
  \caption{Kriging maps of \textsc{when} showing some of the systematic variation in the dataset.}
  \label{fig:variation}
\end{figure*}

The semantic maps for several other varieties from different language families show a division of labor similar to the Huichol one, between lexified means at the bottom of the map and $n$-gram groups (i.e. likely morphologically encoded \textsc{when}-clauses) at the top of the map, as in Chuy (Mayan, Guatemala; Figure \ref{cac}), Comaltepec Chinantec (Otomanguean, Mexico; Figure \ref{cco}), or Terena-Kinikinao-Chane (Arawakan, Bolivia; Figure \ref{ter}).

\paragraph{Beyond switch-reference} The integration of character $n$-grams to the semantic map of \textsc{when} was primarily driven by the aim of capturing morphological means of marking generic temporal subordination, which these examples from Latin American languages indicate as promising, especially in light of the known switch-reference markers captured in the maps. However, as mentioned in Section \ref{sec:intro}, there is great linguistic variation in the Latin American and Caribbean region and the new semantic maps helped capture more than just $n$-gram groups overlapping with the switch-reference markers in Huichol or Quechuan languages. Several languages, for instance, show an inverted pattern to the Huichol one, with a lexicalized means at the top of the map and an $n$-gram area at the bottom, as in Ticuna (Ticuna-Yuri, Western Amazon; Figure \ref{tca}) or Lomeriano-Ignaciano Chiquitano (Chiquitano, Bolivia; Figure \ref{cax}).

Yet others only use one $n$-gram for both the bottom and top areas, as in Tabasco Chontal (Mayan, Mexico; Figure \ref{chf}), or only lexified means, as in San Mateo del Mar Huave (Huavean/Isolate, Mexico; Figure \ref{huv}), Nivaclé (Matacoan, Argentina and Paraguay; Figure \ref{cag}), Kaqchikel (Mayan, Guatemala; Figure \ref{cak}), Guerrero Amuzgo (Otomanguean, Mexico; Figure \ref{amu}), Pichis Ashéninka (Arawakan, Peru; Figure \ref{cpu}), and Chamacoco (Zamucoan, Paraguay; Figure \ref{ceg}).

\section{Conclusion \& Future Work }

\paragraph{Summary and findings} This paper has presented probabilistic semantic maps of \textsc{when}-clauses based on a parallel corpus of New Testament translations in Latin American and Caribbean languages. The rationale behind this study was the observation that \textsc{when}-clauses in the Latin American region are often encoded \textit{morphologically} (exclusively or predominantly so, i.e. in addition to lexified subordinators), which in previous token-based experiments (i.e. based only on full-token correspondences between languages) represented one of the main hurdles for the detection of systematic cross-linguistic variation in the expression of generic temporal subordination.

It built on previous approaches based on correspondences between a source word (English \textit{when}) and character $n$-grams, using association measures to detect meaningful groups of $n$-grams that are likely to represent a particular morphological marker encoding temporal subordination in each target language. The approach has yielded results that are clearly helpful in identifying morphologically-encoded \textsc{when}-clauses in languages where switch-reference markers (same-subject or different-subject marking) are employed to mark a predicate as subordinate to their matrix clause. The identification of groups of $n$-grams as switch-reference markers in some of the languages in the corpus was achieved by consulting descriptive grammars and language-specific typological studies (e.g. on the Quechuan morphological system), but also because of the use of Huichol, a Mexican language with switch-reference morphology, as a point of reference to build a small benchmark and optimize hyperparameters during the generation of the semantic maps.

\paragraph{Future research} Future studies may want to experiment with different $n$-gram sizes and different association measures and Kriging parameters, as well as use languages other than Huichol as benchmarks for the calibration of the Kriging models. Languages showing an opposite pattern to that of Huichol (i.e. a lexified means where Huichol has a morphological means, and vice versa) would particularly benefit from a close-reading evaluation to ascertain whether the method did manage to capture morphologically-expressed \textit{when}-clauses as accurately as their opposite pattern. 

Finally, the semantic dimensions in the maps have not been fully analyzed, and future studies will take a systematic approach to identifying clusters of observations that are frequently co-expressed, whether morphologically or lexically, across the languages of the corpus, and will establish whether such clusters represent cross-linguistically relevant gram types.

\section*{Limitations}

The main limitation of this experiment is that evaluation, including hyperparameter optimization for the Kriging models, was based on one particular language, Huichol, because of the well-studied subordination system and the presence of a lexified subordinator in addition to the widely employed morphological means (switch-reference). Moreover, not only is switch-reference only one of the several attested morphological means to convey generic temporal subordination cross-linguistically, but there are also major differences between switch-reference systems (both in terms of the set of markers available to a language, but also their range of functions). The hyperparameters tuning based on Huichol has likely introduced some bias towards languages that have a similar system (i.e. one lexified counterpart to English \textit{when} alongside switch-reference morphology), potentially obscuring other relevant typological dimensions (e.g. systematic clause-bridging marking).

The $n$-gram approach identifies \textit{groups} of character $n$-grams, but does not yet provide a straightforward way of selecting one particular set of characters as the representative morpheme from a series of potential allomorphs. A tentative solution could be extracting the shortest allomorph, or the allomorph representing the common denominator among all $n$-grams in a set. However, this has not been tested and we have simply numbered each group of $n$-grams while keeping track of what forms each group contains for subsequent easier retrieval and inspection, if needed.

\section*{Acknowledgements}
This work was supported by the Ecosystem Leadership Award under the EPSRC Grant EP/X03870X/1 \& The Alan Turing Institute, particularly the Turing Research Fellowship scheme under that grant.

\bibliography{custom}

\begin{thebibliography}{37}
\expandafter\ifx\csname natexlab\endcsname\relax\def\natexlab#1{#1}\fi

\bibitem[{Asgari and Sch{\"u}tze(2017)}]{superpivot}
Ehsaneddin Asgari and Hinrich Sch{\"u}tze. 2017.
\newblock \href {https://doi.org/10.18653/v1/D17-1011} {Past, present, future: A computational investigation of the typology of tense in 1000 languages}.
\newblock In \emph{Proceedings of the 2017 Conference on Empirical Methods in Natural Language Processing}, pages 113--124, Copenhagen, Denmark. Association for Computational Linguistics.

\bibitem[{Bierge(2017)}]{huicholthesis}
Stefanie~Ramos Bierge. 2017.
\newblock \href {https://scholar.colorado.edu/downloads/8g84mm526} {\emph{{Clause Types and Transitivity in Wix{\'a}rika (Huichol): a Uto-Aztecan Language}}}.
\newblock Ph.D. thesis, University of Colorado.

\bibitem[{Comrie(1982)}]{grammaticalrelhuichol}
Bernard Comrie. 1982.
\newblock \href {https://doi.org/10.1163/9789004368903_007} {Grammatical relations in {Huichol}}.
\newblock In \emph{Studies in Transitivity}, pages 95--115. Brill, Leiden, The Netherlands.

\bibitem[{Comrie(1983)}]{comriehuichol}
Bernard Comrie. 1983.
\newblock \href {https://doi.org/10.1075/tsl.2.04com} {{Switch-reference in Huichol: A typological study}}.
\newblock In John Haiman and Pamela Munro, editors, \emph{Switch Reference and Universal Grammar}, pages 17--38. John Benjamins, Amsterdam/Philadelphia.

\bibitem[{Cristofaro(1998)}]{Cristofaro-1998}
Sonia Cristofaro. 1998.
\newblock \href {https://doi.org/10.1524/stuf.1998.51.1.3} {{Deranking and Balancing in Different Subordination Relations: a Typological Study}}.
\newblock \emph{Sprachtypologie und Universalienforschung}, 51:3--42.

\bibitem[{Cristofaro(2019)}]{cristofarosubordination}
Sonia Cristofaro. 2019.
\newblock \href {https://doi.org/https://doi.org/10.1093/acprof:oso/9780199282005.001.0001} {\emph{{Subordination}}}.
\newblock {O}xford University Press, {O}xford.

\bibitem[{Croft(2002)}]{Croft-2002}
William Croft. 2002.
\newblock \href {https://doi.org/https://doi.org/10.1017/CBO9780511840579} {\emph{{Typology and Universals}}}, 2nd edition.
\newblock Cambridge University Press, Cambridge.

\bibitem[{Croft and Poole(2008)}]{croft-poole-2008}
William Croft and Keith~T. Poole. 2008.
\newblock \href {https://doi.org/10.1515/THLI.2008.001} {{Inferring universals from grammatical variation: Multidimensional scaling for typological analysis}}.
\newblock \emph{Theoretical Linguistics}, 34(1):1--37.

\bibitem[{Dahl and Wälchli(2016)}]{dahl}
Östen Dahl and Bernhard Wälchli. 2016.
\newblock \href {https://doi.org/10.15448/1984-7726.2016.3.25454} {{Perfects and iamitives: two gram types in one grammatical space}}.
\newblock \emph{Letras de Hoje}, pages 325--348.

\bibitem[{Evans(2017)}]{pulargrammar}
Barrie Evans. 2017.
\newblock {A teaching grammar of Pular}.
\newblock Ms.

\bibitem[{Getis(2008)}]{getisautocorr}
Arthur Getis. 2008.
\newblock \href {https://doi.org/https://doi.org/10.1111/j.1538-4632.2008.00727.x} {A history of the concept of spatial autocorrelation: A geographer's perspective}.
\newblock \emph{Geographical Analysis}, 40(3):297--309.

\bibitem[{Guillaume(2008)}]{guillaume2008}
Antoine Guillaume. 2008.
\newblock \href {https://doi.org/https://doi.org/10.1515/9783110211771} {\emph{{A Grammar of Cavineña}}}.
\newblock Mouton de Gruyter, Berlin.

\bibitem[{Guillaume(2011)}]{guillaume2011}
Antoine Guillaume. 2011.
\newblock \href {https://doi.org/https://doi.org/10.1075/tsl.97.05gui} {{Subordinate clauses, switch-reference, and tail-head linkage in Cavineña narratives}}.
\newblock In Pieter~Muysken Rik~van Gijn, Katharina~Haude, editor, \emph{Subordination in Native South American languages}, pages 109--140. John Benjamins, Amsterdam/Philadelphia.

\bibitem[{Haiman and Munro(1983)}]{haimanmunro}
John Haiman and Pamela Munro. 1983.
\newblock \href {https://doi.org/https://doi.org/10.1075/tsl.2} {{Introduction}}.
\newblock In John Haiman and Pamela Munro, editors, \emph{Switch Reference and Universal Grammar}, pages ix--xv. John Benjamins, Amsterdam.

\bibitem[{Hammarstr{\"o}m et~al.(2023)Hammarstr{\"o}m, Forkel, Haspelmath, and Bank}]{glottolog2021}
Harald Hammarstr{\"o}m, Robert Forkel, Martin Haspelmath, and Sebastian Bank. 2023.
\newblock \href {https://doi.org/10.5281/zenodo.8131084} {\emph{Glottolog 4.8}}.
\newblock Max Planck Institute for Evolutionary Anthropology, Leipzig.

\bibitem[{Haug and Pedrazzini(2023)}]{anon1}
Dag Haug and Nilo Pedrazzini. 2023.
\newblock \href {https://doi.org/10.3389/fcomm.2023.1163431} {{The semantic map of \textit{when} and its typological parallels}}.
\newblock \emph{Frontiers in Communication}, 8.

\bibitem[{Honnibal and Montani(2017)}]{spacy2}
Matthew Honnibal and Ines Montani. 2017.
\newblock {spaCy 2}: Natural language understanding with {B}loom embeddings, convolutional neural networks and incremental parsing.

\bibitem[{Junczys-Dowmunt and Szał(2012)}]{symgiza}
Marcin Junczys-Dowmunt and Arkadiusz Szał. 2012.
\newblock \href {http://emjotde.github.io/publications/pdf/mjd2011siis.pdf} {{SyMGiza++: Symmetrized Word Alignment Models for Machine Translation}}.
\newblock In \emph{Security and Intelligent Information Systems (SIIS)}, volume 7053 of \emph{Lecture Notes in Computer Science}, pages 379--390, Warsaw, Poland. Springer.

\bibitem[{Keine(2013)}]{keine}
Stephan Keine. 2013.
\newblock \href {https://doi.org/https://doi.org/10.1007/s11049-013-9194-8} {{Deconstructing switch-reference}}.
\newblock \emph{Natural Language and Linguistic Theory}, 31:767–826.

\bibitem[{Levshina(2019)}]{levshina19}
Natalia Levshina. 2019.
\newblock \href {https://doi.org/https://doi.org/10.1515/lingty-2019-0025} {{Token-based typology and word order entropy: A study based on Universal Dependencies}}.
\newblock \emph{Linguistic Typology}, 23(3):533--572.

\bibitem[{Levshina(2022)}]{levshina21}
Natalia Levshina. 2022.
\newblock \href {https://doi.org/10.1515/lingty-2020-0118} {{Corpus-based typology: applications, challenges and some solutions}}.
\newblock \emph{Linguistic Typology}, 26(1):129--160.

\bibitem[{Mayer and Cysouw(2014)}]{mayer-cysouw}
Thomas Mayer and Michael Cysouw. 2014.
\newblock \href {http://www.lrec-conf.org/proceedings/lrec2014/pdf/220_Paper.pdf} {{Creating a massively parallel Bible corpus}}.
\newblock In \emph{Proceedings of the Ninth International Conference on Language Resources and Evaluation ({LREC}'14)}, pages 3158--3163, Reykjavik, Iceland. European Language Resources Association (ELRA).

\bibitem[{McKenzie(2012)}]{mckenzie2012}
Andrew McKenzie. 2012.
\newblock \href {https://doi.org/https://doi.org/10.7275/tfp7-0s60} {\emph{{The role of contextual restriction in reference-tracking}}}.
\newblock Ph.D. thesis, Amherst: University of Massachusetts.

\bibitem[{McKenzie(2015{\natexlab{a}})}]{mckenzie2015a}
Andrew McKenzie. 2015{\natexlab{a}}.
\newblock \href {https://doi.org/https://doi.org/10.1086/681580} {{A survey of switch-reference in North America}}.
\newblock \emph{International Journal of American Linguistics}, 81:409–448.

\bibitem[{McKenzie(2015{\natexlab{b}})}]{mckenzie2015b}
Andrew McKenzie. 2015{\natexlab{b}}.
\newblock {Austinian situations and switch-reference: The role of context in reference-tracking.}
\newblock Ms.

\bibitem[{Müller et~al.(2023)Müller, Yurchak, Murphy, nannau, Ziebarth, Basak, Albuquerque, Vrijlandt, Peveler, Raigosa, Matchette-Downes, Porter, Rhilip, Staniewicz, Chang, and kvanlombeek}]{pykrige}
Sebastian Müller, Roman Yurchak, Benjamin Murphy, nannau, Malte Ziebarth, Sudipta Basak, Marcelo Albuquerque, Mark Vrijlandt, Matthew Peveler, Daniel~Mejía Raigosa, Harry Matchette-Downes, Jordan Porter, Rhilip, Scott Staniewicz, Will Chang, and kvanlombeek. 2023.
\newblock \href {https://doi.org/10.5281/zenodo.10016909} {Geostat-framework/pykrige: v1.7.1}.

\bibitem[{Nordhoff and Hammarström(2011)}]{nordhoffhammarstrom}
Sebastian Nordhoff and Harald Hammarström. 2011.
\newblock \href {http://iswc2011.semanticweb.org/fileadmin/iswc/Papers/Workshops/LISC/nordhoff.pdf} {{Glottolog/Langdoc: Defining dialects, languages, and language families as collections of resources}}.
\newblock In \emph{Proceedings of ISWC 2011}.

\bibitem[{Overall(2014)}]{overall2014}
Simon~E. Overall. 2014.
\newblock \href {https://doi.org/https://doi.org/10.1075/tsl.105.11ove} {{Clause-chaining, switch-reference and nominalisations in Aguaruna (Jivaroan)}}.
\newblock In Rik van Gijn, Jeremy Hammond, Dejan Matić, Saskia van Putten, and Ana Vilacy~Galucio, editors, \emph{Information Structure and Reference Tracking in Complex Sentences}, pages 309--340. John Benjamins, Amsterdam/Philadelphia.

\bibitem[{Overall(2016)}]{overall2016}
Simon~E. Overall. 2016.
\newblock \href {https://doi.org/https://doi.org/10.1075/tsl.114.13ove} {{Switch-reference and case-marking in Aguaruna (Jivaroan) and beyond}}.
\newblock In Rik van Gijn and Jeremy Hammond, editors, \emph{Switch Reference 2.0}, page 453–472. John Benjamins, Amsterdam/Philadelphia.

\bibitem[{Pedrazzini(2023)}]{anon2}
Nilo Pedrazzini. 2023.
\newblock \href {https://doi.org/10.5287/ora-8gv0b4qyo} {\emph{A quantitative and typological study of Early Slavic participle clauses and their competition}}.
\newblock Ph.D. thesis, University of Oxford.

\bibitem[{Roberts(2017)}]{roberts_2017}
John~R. Roberts. 2017.
\newblock \href {https://doi.org/10.1017/9781316135716.017} {{A Typology of Switch Reference}}.
\newblock In Alexandra~Y. Aikhenvald and R.~M.~W.Editors Dixon, editors, \emph{The Cambridge Handbook of Linguistic Typology}, Cambridge Handbooks in Language and Linguistics, page 538–573. Cambridge University Press.

\bibitem[{Stassen(1985)}]{Stassen-1985}
Leon Stassen. 1985.
\newblock \href {https://www.degruyter.com/database/COGBIB/entry/cogbib.11332/html} {\emph{{Comparison and Universal Grammar}}}.
\newblock Basil Blackwell, Oxford.

\bibitem[{Stirling(1993)}]{stirling_1993}
Lesley Stirling. 1993.
\newblock \href {https://doi.org/10.1017/CBO9780511597886} {\emph{{Switch-Reference and Discourse Representation}}}.
\newblock Cambridge University Press.

\bibitem[{van Gijn(2012)}]{vangijn2012southame}
Rik van Gijn. 2012.
\newblock \href {https://doi.org/10.1349/PS1.1537-0852.A.407} {{Switch-attention (aka switch-reference) in South-American temporal clauses}}.
\newblock \emph{Linguistic Discovery}, 10(1):112--27.

\bibitem[{van Gijn(2016)}]{vangijn2016srsouthame}
Rik van Gijn. 2016.
\newblock \href {https://doi.org/https://doi.org/10.1075/tsl.114.05van} {{Switch reference in Western South America}}.
\newblock In Rik van Gijn and Jeremy Hammond, editors, \emph{Switch Reference 2.0}, page 153–206. John Benjamins, Amsterdam/Philadelphia.

\bibitem[{van Gijn et~al.(2011)van Gijn, Haude, and Muysken}]{gijnetalsouthame}
Rik van Gijn, Katharina Haude, and Pieter Muysken, editors. 2011.
\newblock \href {https://doi.org/https://doi.org/10.1075/tsl.97} {\emph{{Subordination in native South American languages}}}.
\newblock John Benjamins, Amsterdam/Philadelphia.

\bibitem[{Wälchli and Cysouw(2012)}]{walchli-cysouw2012}
Bernhard Wälchli and Michael Cysouw. 2012.
\newblock \href {https://doi.org/10.1515/ling-2012-0021} {{Lexical typology through similarity semantics: Toward a semantic map of motion verbs}}.
\newblock \emph{Linguistics}, 50:671--710.

\end{thebibliography}

\appendix

\end{document}